\documentclass[lettersize,journal]{IEEEtran}
\usepackage{amsmath,amsfonts}
\usepackage{algorithm}
\usepackage{array}
\usepackage[caption=false,font=normalsize,labelfont=sf,textfont=sf]{subfig}
\usepackage{textcomp}
\usepackage{url}
\usepackage{verbatim}
\usepackage{graphicx}
\usepackage{tabularx}
\usepackage{multirow}    
\usepackage{rotating}   
\usepackage[table]{xcolor}
\usepackage{booktabs}
\usepackage{amsmath}
\usepackage{amssymb}
\usepackage{algpseudocode}
\usepackage{bm}
\usepackage{placeins}
\usepackage[colorlinks=true]{hyperref}
\hypersetup{
	citecolor = green,
	linkcolor = red,
	urlcolor  = magenta
}

\begin{document}

\title{FeedbackSTS-Det: Sparse Frames-Based Spatio-Temporal Semantic Feedback Network for Moving Infrared Small Target Detection}

\author{Yian Huang,~\IEEEmembership{Student Member,~IEEE,}~
	~Qing Qin$^{*}$,
	Aji Mao,~
	Xiangyu Qiu,~\IEEEmembership{Student Member,~IEEE,}~
	Liang Xu,~
	Xian Zhang,~
	Zhenming Peng$^{*}$,~\IEEEmembership{Senior Member,~IEEE}
\thanks{This work was supported by the Natural Science Foundation of Sichuan Province of China (Grant No.2025ZNSFSC0522) and partially supported by the National Natural Science Foundation of China (Grant No.61571096). (Corresponding author: Zhenming Peng, Qing Qin).}
\thanks{Yian Huang, Aji Mao, Xiangyu Qiu, Liang Xu, Xian Zhang, and Zhenming Peng are with the School of Information and Communication Engineering and the Laboratory of Imaging Detection and Intelligent Perception, University of Electronic Science and Technology of China, Chengdu 611731, China~(e-mail: huangyian1@std.uestc.edu.cn; ajimao@std.uestc.edu.cn; xiangyuqiu@std.uestc.edu.cn; liangx@std.uestc.edu.cn; zhangxian@std.uestc.edu.cn; zmpeng@uestc.edu.cn).\\
Qing Qin is with School of Television and Audiovisual Arts, Communication University of Zhejiang, Hangzhou 310018, China~(e-mail: 20190012@cuz.edu.cn).}
}

\markboth{IEEE Transactions on Circuits and Systems for Video Technology}%
{Huang \MakeLowercase{\textit{et al.}}: FeedbackSTS-Det for Moving Infrared Small Target Detection}


\maketitle

\begin{abstract}
Infrared small target detection (ISTD) has been a critical technology in defense and civilian applications over the past several decades, such as missile warning, maritime surveillance, and disaster monitoring. Nevertheless, moving infrared small target detection still faces considerable challenges: existing models suffer from insufficient spatio-temporal semantic correlation and are not lightweight-friendly, while algorithms with strong scene generalization capability are in great demand for real-world applications. To address these issues, we propose FeedbackSTS-Det, a sparse frames-based spatio-temporal semantic feedback network. Our approach introduces a closed-loop spatio-temporal semantic feedback strategy with paired forward and backward refinement modules that work cooperatively across the encoder and decoder to enhance information exchange between consecutive frames, effectively improving detection accuracy and reducing false alarms. Moreover, we introduce an embedded sparse semantic module (SSM), which operates by strategically grouping frames by interval, propagating semantics within each group, and reassembling the sequence to efficiently capture long-range temporal dependencies with low computational overhead. Extensive experiments on many widely adopted multi-frame infrared small target datasets demonstrate the generalization ability and scene adaptability of our proposed network. Code and models are available at: \url{https://github.com/IDIP-Lab/FeedbackSTS-Det}.   
\end{abstract}


\section{Introduction}
\IEEEPARstart{I}{nfrared} small target detection (ISTD) has served as a fundamental and critical technology over the past several decades, playing an essential role in both defense and civilian applications such as missile warning \cite{zhang2022learning, ying2022local}, maritime surveillance \cite{teutsch2010classification, hu2024smpisd}, and disaster monitoring \cite{tramutoli2015visual}. Although traditional single-frame infrared small target (SIRST) detection methods have achieved notable progress, they suffer from an inherent limitation: the inability to leverage temporal motion cues, resulting in insufficient robustness under low signal-to-noise ratio (SNR) scenarios. To address this bottleneck, multi-frame infrared small target detection (MIRSTD) has emerged, which significantly enhances detection performance by integrating temporal contextual information.

Existing MIRST detection approaches can be broadly categorized into two technical paradigms. The first is model-driven methods, which rely on mathematical priors for target-background separation. Early filtering-based approaches \cite{tom1993morphology, hadhoud1988two}, and subsequent optimization-based methods leveraging low-rank and sparse decomposition \cite{gao2013infrared, dai2017reweighted, zhang2018infrared} have demonstrated effectiveness in specific scenarios. More recent tensor-based extensions incorporate temporal information through spatio-temporal modeling \cite{sun2020infrared, wang2021infrared, liu2023infrared}. Although theoretically principled, these methods often require meticulous manual tuning and struggle to adapt to complex, non-linear clutter environments.

The second category comprises data-driven methods, which employ deep learning to automatically learn spatio-temporal features. Within single-frame detection, substantial efforts have focused on enhancing spatial feature discrimination through multi-scale fusion \cite{dai2021asymmetric, zhang2023attention}, dense connections \cite{li2022dense, wu2022uiu}, and feature grouping mechanisms \cite{hou2022istdunet}. Another important direction incorporates mathematical priors into network architectures to improve interpretability and robustness \cite{zhang2022isnet, sun2023receptive, wu2024saliency, hu2025datransnet}, while deep unfolding networks \cite{wu2024rpcanet, liu2025ctvnet, xiong2025drpca} further bridge model-driven and data-driven paradigms. For multi-frame detection, temporal information is leveraged through various strategies, including unsupervised sequence registration \cite{hou2024unsupervised}, motion modeling \cite{li2023direction}, deformable alignment \cite{ying2025infrared}, nonlinear spatio-temporal feature enhancement \cite{wu2026neural}, and temporal anomaly reformulation \cite{li2025probing}. Despite their advancements, data-driven methods face the following challenges:

Infrared small targets typically occupy only a few pixels, lack distinct shape or texture features, and are easily hidden in complex backgrounds. Without sufficient use of temporal information, effective target detection becomes very difficult. Nevertheless, excessively long sequences greatly increase computational cost and make lightweight design difficult to achieve. In addition, abundant false alarm sources and background clutter often make algorithm performance unstable. Therefore, infrared small target detection algorithms require extremely high scene adaptability.

To tackle these challenges, this paper proposes a novel sparse frames-based spatio-temporal feedback semantics network, termed FeedbackSTS-Det. The core of our approach is a spatio-temporal semantic feedback strategy that establishes a closed-loop semantic association mechanism. Specifically, it integrates a forward spatio-temporal semantic refinement module (FSTSRM) in the encoder and a backward spatio-temporal semantic refinement module (BSTSRM) in the decoder in a cooperative manner, both built upon a balanced 3D Res-UNet backbone. Additionally , we embed a sparse semantic module (SSM) within both refinement modules, which performs structured sparse temporal modeling to capture long-range dependencies with low computational cost. These designs facilitate robust implicit inter-frame registration and continuous spatio-temporal semantic refinement, ensuring effective false alarm suppression and consistent performance. Furthermore, extensive experiments on many widely adopted multi-frame infrared small target datasets demonstrate the generalization ability and scene adaptability of our proposed network.

The main contributions of this work are summarized as follows:

\begin{itemize}
	\item{We propose a closed-loop semantic feedback strategy for infrared small target detection, comprising paired forward and backward refinement modules that operate across the encoder and decoder. This design enhances information exchange between consecutive frames, effectively improving detection accuracy and reducing false alarms.}
	\item{We introduce an embedded sparse semantic module (SSM), which operates by strategically grouping frames by interval, propagating semantics within each group, and reassembling the sequence to efficiently capture long-range temporal dependencies with low computational overhead.}
	\item{Our model exhibits strong detection capability and false alarm suppression across all experiments on many widely adopted multi-frame infrared small target datasets, demonstrating its generalization ability and scene adaptability.}
\end{itemize}

The remainder of this paper is organized as follows: Section \ref{sec:related_work} provides a comprehensive review and critical analysis of related work. Section \ref{sec:methodology} describes the proposed network architecture. Section \ref{sec:experiment} presents experimental results and analysis. Finally, Section \ref{sec:conclusion} provides concluding remarks.

\section{RELATED WORK} \label{sec:related_work}
In this section, we provide a concise review of the main methods in the field of IRST, with a particular emphasis on research related to MIRST detection. Existing MIRST detection methods can be broadly categorized into model-driven methods and data-driven methods. The former primarily relies on filtering or handcrafted optimization models for feature extraction, while the latter utilizes deep learning methods to automatically learn feature representations from large-scale data.

\subsection{Model-driven schemes for IRST}
Model-driven infrared small target detection methods can be divided into two categories: filtering methods based on local priors and optimization methods based on non-local priors.

Filtering methods operate under the background consistency assumption, estimating the image background and enhancing targets via background subtraction. Classical methods like Top-Hat \cite{tom1993morphology} and two-dimensional least mean square(TDLMS) \cite{hadhoud1988two} perform well in uniform backgrounds but struggle in complex cluttered environments. Subsequent improvements include methods leveraging the human visual system (HVS) to exploit target saliency characteristics \cite{kim2012scale, chen2013local, wei2016multiscale}.

In contrast, optimization methods exploit the non-local correlation of backgrounds and target sparsity, achieving target-background separation through model formulation. Assuming low-rank background subspaces, Gao et al. \cite{gao2013infrared} models ISTD as a low-rank and sparse decomposition problem. This inspired numerous subspace-based variants \cite{he2015small, wang2017infrared, zhang2021infrared}. Further developments include designing the reweighted infrared patch tensor \cite{dai2017reweighted}, formulating the non-convex rank approximation minimization\cite{zhang2018infrared}, proposing the partial sum of tensor nuclear norm\cite{zhang2019infrared}, and incorporating local contrast energy or hyper total variation to improve robustness \cite{guan2020infrared, kong2021infrared}.

To incorporate temporal information, sequence-based tensor methods have been developed, extending prior spatial models into the spatio-temporal domain. Representative approaches include extending multi-subspace learning to the tensor domain \cite{sun2020infrared}, constructing spatio-temporal tensors with overlapping patches \cite{liu2020small}, and extracting non-overlapping patches across frames via sliding windows \cite{wang2021infrared}. Further improvements involve designing specialized regularization methods combined with non-convex tensor rank approximation \cite{liu2021nonconvex}, imposing constraints on Tucker decomposition factor matrices \cite{liu2023infrared}, and incorporating total variation on sparse representation matrices \cite{liu2023representative}. Advanced methods also leverage sparse regularization to enhance target-background distinction \cite{li2023sparse} or extend tensor decomposition to four-dimensional representations \cite{wu2023infrared}.
\subsection{Data-driven schemes for IRST}
Model-driven methods often rely on manual parameter tuning and idealized mathematical models, whereas data-driven methods employ deep learning to automatically learn features from large-scale data, enabling end-to-end infrared small target detection.

Within this data-driven paradigm, single-frame detection focuses on extracting more discriminative spatial features from a single image. Specifically, Dai et al. \cite{dai2021asymmetric} and Zhang et al. \cite{zhang2023attention} enhance feature representation for small targets by designing multi-scale asymmetric fusion modules that synergize global context attention with local attention mechanisms targeting point-like objects. Similarly, Li et al. \cite{li2022dense} and Wu et al. \cite{wu2022uiu} strengthen internal connections among network modules and the flow of feature information through repeated stacking of model structures. Hou et al. \cite{hou2022istdunet} introduce a feature map grouping mechanism and assign higher weights to groups containing small targets to focus on critical information. 

Beyond these purely data-driven designs, another important direction is to incorporate mathematical model priors into neural network architectures. For instance, Zhang et al. \cite{zhang2022isnet} design an edge region extraction module by mimicking the analytical process of neural ordinary differential equations. Sun et al. \cite{sun2023receptive} and Zhang et al. \cite{zhang2024mdigcnet} embed Gaussian operator modules into networks based on the assumption that the grayscale distribution of infrared targets approximates a Gaussian distribution.  Meanwhile, research on improving model interpretability has also gained attention. For example, Wu et al. \cite{wu2024saliency} design learnable local saliency kernels to embed traditional filtering principles into networks, while Hu et al. \cite{hu2025datransnet} develop a specific convolution block that simulates central difference to extract image gradient features more robustly. 

A notable advancement in bridging model-driven and data-driven methodologies is the adoption of deep unfolding networks (DUNs). Built upon on robust principal component analysis \cite{goldfarb2014robust}, RPCANet \cite{wu2024rpcanet} demonstrates the feasibility of applying DUNs to ISTD. Liu et al. \cite{liu2025ctvnet, liu2025ddfet} further introduce background gradient priors into DUNs to improve the accuracy of background estimation. Subsequently, Xiong et al. \cite{xiong2025drpca} construct a lightweight dynamic unfolding mechanism to enhance model robustness and generalization across diverse backgrounds within the framework of DUNs. Furthermore, Liu et al. \cite{deng2025dusrnet} embed bottleneck layers into DUNs to reduce the number of channels during feature extraction and incorporate denoising modules to strengthen robustness against complex noise.

In contrast to single-frame approaches, MIRSTD methods place greater emphasis on mining and utilizing temporal information across consecutive frames. Hou et al. \cite{hou2024unsupervised} propose a hierarchical temporal consistency loss function within an unsupervised end-to-end framework to resolve registration inconsistencies of infrared image sequences. Li et al. \cite{li2023direction} extract spatial saliency features via a detection head, then input the features of five consecutive frames into a self-designed direction-coded temporal U-shape module (DTUM) for motion modeling. The recurrent feature refinement framework (RFR) \cite{ying2025infrared} employs deformable convolutions and a pyramid structure to enhance inter-frame registration and positional information extraction for long-sequence detection. Wu et al. \cite{wu2026neural} further employs nonlinear networks to enhance correlations between spatial-temporal features and performs detection in an unsupervised manner. Furthermore, Li et al. \cite{li2025probing} reformulate infrared small target detection as a one-dimensional signal anomaly detection problem, where global temporal saliency and correlation characteristics distinguish targets from interference signals.

\section{METHODOLOGY} \label{sec:methodology}

\begin{figure*}[!t]
    \centering
    \includegraphics[width=1.0\textwidth]{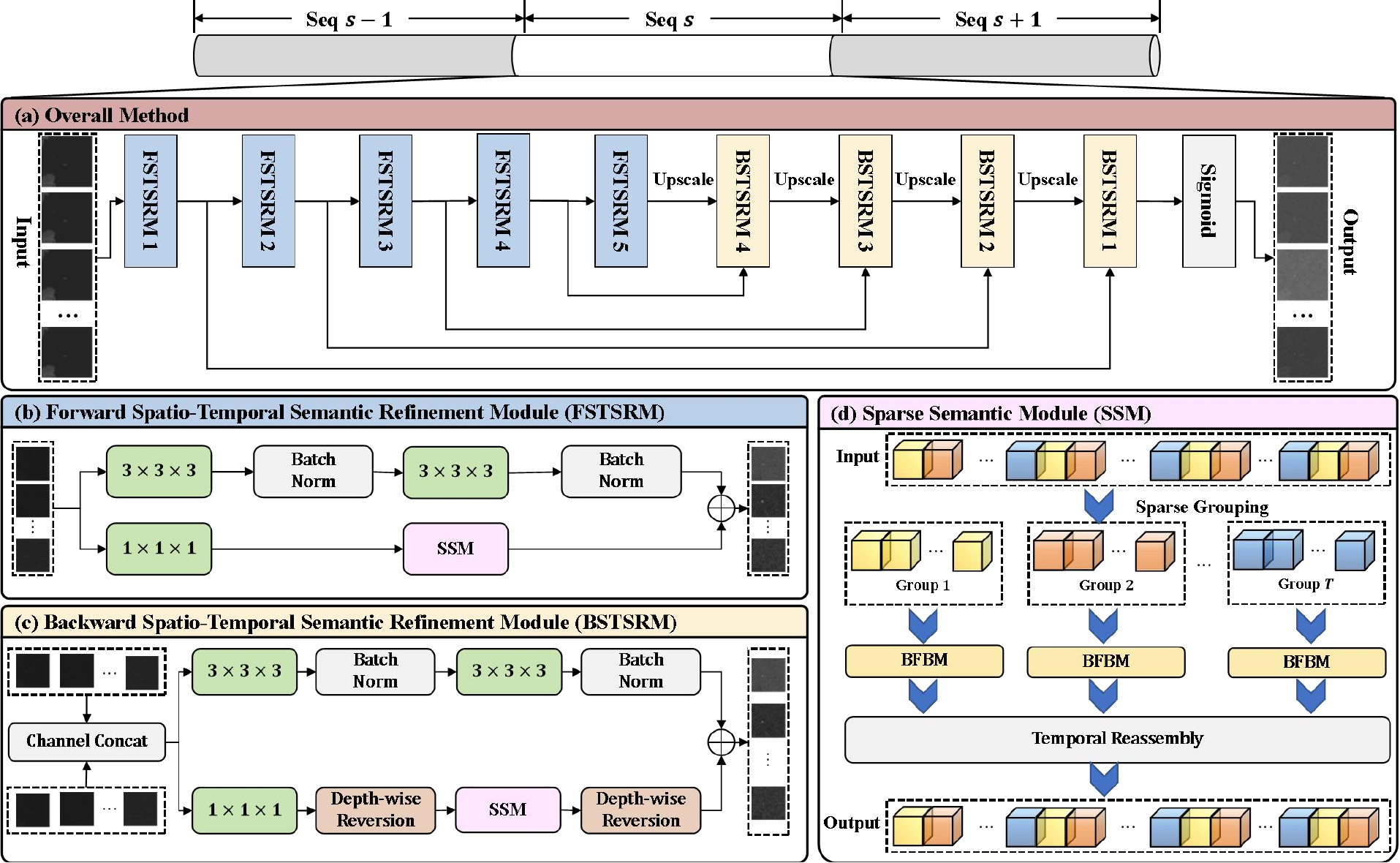}
    \caption{Overall procedure of FeedbackSTS-Det model for ISTD. (a) presents the overall framework. (b) denotes forward spatio-temporal semantic refinement module (FSTSRM), (c) illustrates backward spatio-temporal semantic refinement module (BSTSRM), and (d) describes sparse semantic module (SSM). }
    \label{fig:proposed_framework}
\end{figure*}
\subsection{Overall Framework}
As illustrated in Fig.~\ref{fig:proposed_framework}, we present a sparse frames-based spatio-temporal feedback semantic network (FeedbackSTS-Det), which fundamentally rethinks temporal modeling in infrared small target detection. Built upon a balanced 3D Res-UNet backbone~\cite{he2016deep,cciccek20163d}, our network processes sliding windows by partitioning the continuous temporal image data into shorter sequences (denoted as $Seq$) of length $D$, with each $Seq$ sequentially extracted as input. This approach maintains consistency between the training and inference processes, ensuring the stability of model performance during actual deployment.

The encoder phase incorporates five consecutive forward spatio-temporal semantic refinement modules (FSTSRM), while the decoder phase employs four successive backward spatio-temporal semantic refinement modules (BSTSRM). Both modules critically integrate our novel sparse semantic module (SSM), which operates by strategically grouping frames at intervals, propagating semantics within each group using the Basic Feedback Module (BFM), and then reassembling the sequence. Collectively, FSTSRM and BSTSRM establish a spatio-temporal semantic feedback strategy.

The spatio-temporal semantic feedback strategy will be detailed in Section~\ref{sec:semantic_feedback_framework}, whereas the SSM will be elaborated in Section~\ref{sec:sparse_semantic_module}. To alleviate the excessive growth of floating-point operations during iterative feature propagation, we deliberately set the base channel number of the overall framework to $8$, ensuring that the deepest layer maintains a maximum channel count of $128$.
\subsection{Spatio-Temporal Semantic Feedback Strategy}\label{sec:semantic_feedback_framework}
The spatio-temporal semantic feedback strategy establishes a closed-loop semantic association mechanism, which comprises the FSTSRM and BSTSRM described in detail below.

\subsubsection{Forward Spatio-Temporal Semantic Refinement Module}
The forward spatio-temporal semantic refinement module (FSTSRM) establishes the forward propagation path in our spatio-temporal feedback system, progressively building spatio-temporal semantic understanding across frames.  Five consecutive FSTSRMs constitute the decoding layer of the overall framework.

Let $\mathbf{X}^{(l)} \in \mathbb{R}^{C \times D \times H \times W}$ denote the input feature map of the encoding layer at layer $l$. As illustrated in Fig.~\ref{fig:proposed_framework}(b),  forward refinement can be expressed as a bidirectional information integration process:
\begin{equation}
	\mathbf{Y}_{\text{FSTSRM}}^{(l)} = \mathcal{B}_{\text{context}}(\mathbf{X}^{(l)}) \oplus \mathcal{P}_{\text{forward}}(\mathbf{X}^{(l)})
	\label{eq:fstsr_decomp}
\end{equation}
where $\oplus$ denotes element-wise addition. Specifically, $\mathcal{B}{\text{context}}$ consists of two consecutive 3D convolutions with batch normalization to preserve local spatio-temporal structures, and its formulation is given in Eq.~\ref{eq:spatial_context}. Meanwhile, the forward temporal propagation path $\mathcal{P}{\text{forward}}$ employs a sparse semantic modeling strategy, which is defined in Eq.~\ref{eq:forward_time}.
\begin{equation}
	\mathcal{B}_{\text{context}}(\mathbf{X}) = \text{BN}(\text{Conv}_{3\times3\times3}(\text{BN}(\text{Conv}_{3\times3\times3}(\mathbf{X}))))
	\label{eq:spatial_context}
\end{equation}
\begin{equation}
	\mathcal{P}_{\text{forward}}(\mathbf{X}) = \text{SSM}(\text{Conv}_{1\times1\times1}(\mathbf{X}))
	\label{eq:forward_time}
\end{equation}
where the operator $\text{SSM}(\cdot)$ represents the processing through the SSM, whose details are provided in Fig.~\ref{fig:proposed_framework}(d) and Section~\ref{sec:sparse_semantic_module}. 

The design of FSTSRM enables complementary information integration. The context branch provides high-fidelity spatial cues, while the sparse propagation branch establishes long-range temporal associations. 
\subsubsection{Backward Spatio-Temporal Semantic Refinement Module}
The backward spatio-temporal semantic refinement module (BSTSRM) implements retrospective refinement of FSTSRM by reversing temporal dependencies. Four consecutive BSTSRMs modules constitute the decoding layer of the overall framework. 

Let $\mathbf{X}_{\text{dec}}^{(l)} \in \mathbb{R}^{C \times D \times H \times W}$ denote the input feature map of the $l$-th decoding layer, where $l \in [1,5]$ and larger $l$ indicates deeper layers. As illustrated in Fig.~\ref{fig:proposed_framework}(a) and Fig.~\ref{fig:proposed_framework}(c), the input of the BSTSRM at layer $5$ is formed by concatenating the outputs of the two deepest FSTSRM along the channel dimension (Eq.~\ref{eq:decoder_bottom}). For $l \in [1,4]$, the input concatenates the output of the BSTSRM from the deeper layer with the output of the FSTSRM at the same level along the channel dimension (Eq.~\ref{eq:decoder_subsequent}).

\begin{equation}
	\mathbf{X}_{\text{dec}}^{(5)} = \text{Concat}\left(\mathbf{Y}_{\text{FSTSRM}}^{(4)},\ \text{Up}\left(\mathbf{Y}_{\text{FSTSRM}}^{(5)}\right)\right)
	\label{eq:decoder_bottom}
\end{equation}
\begin{equation}
	\mathbf{X}_{\text{dec}}^{(l)} = \text{Concat}\left(\mathbf{Y}_{\text{FSTSRM}}^{(l)},\ \text{Up}\left(\mathbf{Y}_{\text{BSTSRM}}^{(l+1)}\right)\right)
	\label{eq:decoder_subsequent}
\end{equation}

where $\text{Up}(\cdot)$ denotes upsampling.

Building upon this input, the backward refinement process is then expressed as:
\begin{equation}
	\mathbf{Y}_{\text{BSTSRM}}^{(l)} = \mathcal{B}_{\text{context}}(\mathbf{X}_{\text{dec}}^{(l)}) \oplus \mathcal{P}_{\text{backward}}(\mathbf{X_{\text{dec}}}^{(l)})
	\label{eq:bstsr_decomp}
\end{equation}
where the backward feature refinement path $\mathcal{P}{\text{backward}}$ is further defined as:
\begin{equation}
	\mathcal{P}_{\text{backwad}}(\mathbf{X}) = \mathcal{R}\left( \text{SSM}\left( \mathcal{R}\left( \text{Conv}_{1\times1\times1}(\mathbf{X}) \right) \right) \right)	
\end{equation}
where the depth-wise reversal operator $\mathcal{R}$ is the core operation of BSTSRM. Given an input tensor $\mathbf{X}\in \mathbb{R}^{C \times D \times H \times W}$ , $\mathcal{R}$ is defined as:
\begin{equation}
	\mathcal{R}(\mathbf{X}_{c,d,h,w}) = \mathbf{X}_{c, D-d+1, h, w}, \quad \forall d \in [1, D]
\end{equation}

\subsubsection{Feedback Strategy}
As illustrated in Fig.~\ref{fig:proposed_framework}(a), five consecutive FSTSRMs progressively extract spatio-temporal semantic features from low to high levels. By simultaneously preserving spatial context information and performing forward sparse propagation, they enable forward information flow, thereby constituting the feedforward pathway. Meanwhile, four consecutive BSTSRMs reverse the temporal dimension, allowing deep-layer spatio-temporal semantic information to propagate backward to shallower layers, thus forming the feedback pathway. Throughout the propagation process, continuous information exchange between the two pathways facilitates the overall \emph{spatio-temporal semantic feedback strategy}. This feedback structure not only inherits the advantages of traditional networks by leveraging preceding frames to influence the detection results of subsequent frames, but also allows subsequent frames to correct errors in preceding ones, thereby enhancing system robustness, reducing false alarm rates, and improving detection accuracy.

\subsection{Sparse Semantic Module}\label{sec:sparse_semantic_module}

\begin{figure}[!t]
	\centering
	\subfloat[Basic Feedback Module (BFBM)]{%
		\includegraphics[width=\linewidth]{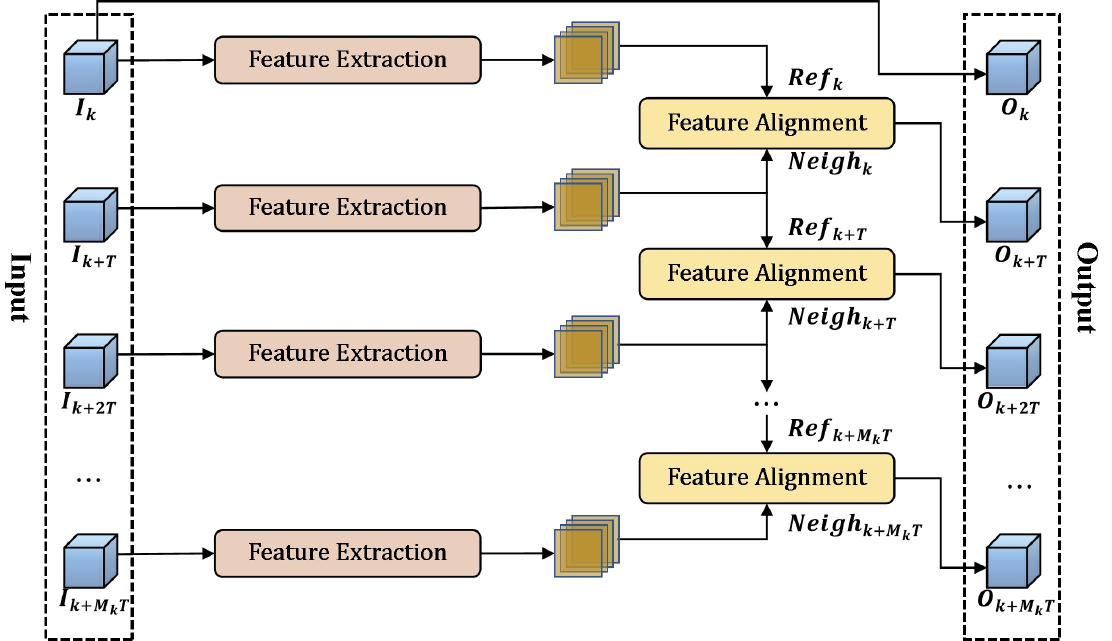}%
		\label{fig:fbm_overll_architecture}%
	}%
	\\
	\subfloat[The Procedure of Feature Extraction]{%
		\includegraphics[width=\linewidth]{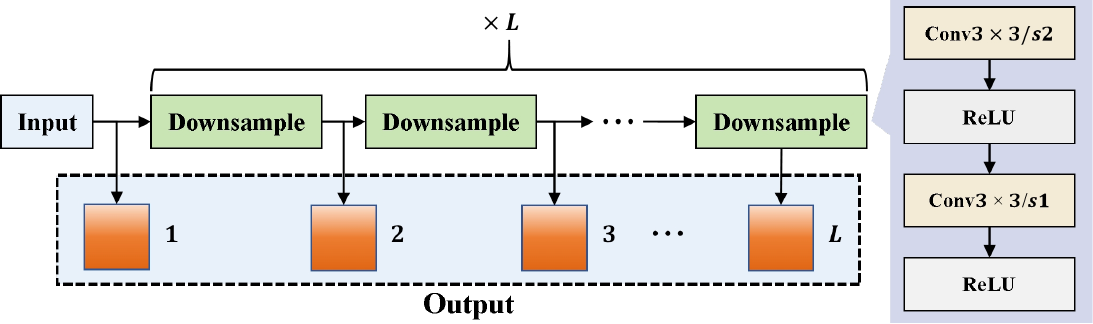}%
		\label{fig:fbm_feature_extraction}%
	}%
	\\
	\subfloat[The Procedure of Feature Alignment]{%
		\includegraphics[width=\linewidth]{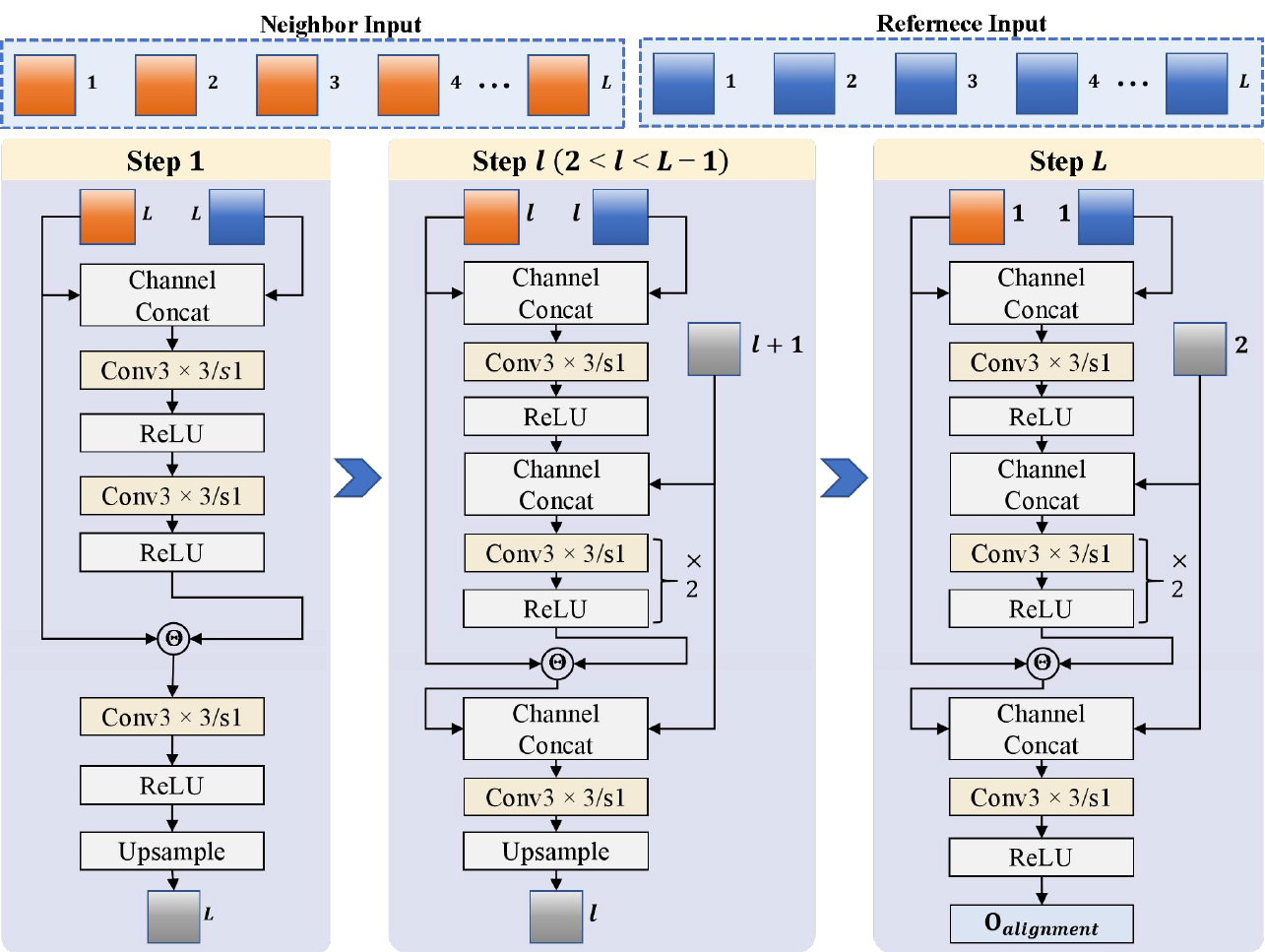}%
		\label{fig:fbm_feature_alginment}%
	}%
	\caption{Basic feedback module (BFBM). (a) presents the framework of the BFBM module, which is embedded within the SSM. (b) details the procedure of feature extraction in BFBM, and (c) shows the procedure of feature alignment module in BFBM.}
	\label{fig:fbm}
\end{figure}

The sparse semantic module (SSM) is incorporated into the temporal feature extraction paths of both the FSTSRM and BSTSRM modules, with its specific structure illustrated in Fig.~\ref{fig:proposed_framework} (d). It is designed to achieve efficient long‑range temporal modeling by strategically processing a subset of frames, thereby reducing computational overhead while enhancing feature propagation across distant frames. The SSM operates in three consecutive stages: sparse grouping, intra‑group spatio‑temporal semantic propagation, and temporal reassembly.

\subsubsection{Sparse Grouping}
Given a sampling interval $T \in \mathbb{N}^+$ and an input feature sequence ${I_1, I_2, \dots, I_{D}}$ of length $D$, the sparse grouping partitions the index set $\mathcal{D} = {1, 2, \dots, D}$ into $T$ disjoint subsequences:
\begin{equation}
	\mathcal{G}_k = \left\{I_i\mid i\in\mathcal{D}, i \equiv k \ (\text{mod}\ T)\right\}, \quad k = 1, 2, \dots, T
	\label{eq:group_define}
\end{equation}

Equivalently, the $k$-th group contains frames with indices $\{k, k+T, k+2T, \dots, k+M_kT\}$, where $M_k = \lfloor (D-k)/T \rfloor$. This constructed index table is also of great help for reorganizing the grouped input feature groups.

In our model framework, there are five encoder layers and four decoder layers. During the feature propagation between frames at the same network layer, the scale of the feature maps remains unchanged, and the number of feature propagation operations is also \(D - T\). Therefore, the computational cost is consistent across these operations. Let \(M_i\) denote the computational cost of feature propagation between frames at the \(i\)-th layer. Then, the overall computational cost of the network is given as follows:
\begin{equation}
	\mathcal{T}\left(T\right)=\left(\sum_{i=1}^{4} 2M_i+M5\right)\left(D-T\right)
	\label{eq:complexity}
\end{equation}
When $T=1$, the model reduces to a sequential model without the sparse mechanism. The reduction in computational cost achieved by the sparse mechanism is given by Eq.~\ref{eq:complexity_reduction}. It can be observed that as the sparse interval $T$ increases, the degree of computational cost reduction becomes larger.
\begin{equation}
	\mathcal{T}_{\text{red}}\left(T\right)=\mathcal{T}\left(1\right)-\mathcal{T}\left(T\right)=\left(\sum_{i=1}^{4} 2M_i+M5\right)(T-1)
	\label{eq:complexity_reduction}
\end{equation}
\subsubsection{Intra-Group Spatio-Temporal Semantic Propagation}
Although the sparse mechanism is an effective approach for reducing computational cost, simply decreasing the number of propagation steps inevitably undermines the effectiveness of feature propagation. Therefore, it is crucial to strengthens the propagation ability within each group. To address this, we propose the basic feedback module (BFBM), as illustrated in the Fig.~\ref{fig:fbm} (a). For each group, all frames except the first one undergo two procedures which are feature extraction (illustrated in Fig.~\ref{fig:fbm} (b)) and feature alignment (illustrated in Fig.~\ref{fig:fbm} (c)). Both procedures are adapted from the pyramid deformable alignment (PDA) module~\cite{ying2025infrared} and are deliberately decoupled in our design to reduce redundant computation. 

Specifically, for a given feature group $\mathcal{G}_k$ obtained according to Eq.~\ref{eq:group_define}, the initial frame $I_{k}$ serves only as a reference. Consequently, the first output $O_k$ within the group is obtained by directly passing $I_{k}$ through without any special processing, as formulated in Eq.~\ref{eq:ssm_init_feat}. For the remaining outputs $O_{k+mt}$, where $m \in \left[1, M \right] \cap \mathbb{N}$, they are generated by first extracting features from the current input frame, which serves as the neighbor input, and its adjacent counterpart, which serves as the reference input. This is followed by the procedure of feature alignment. The corresponding computational process is given by Eq.~\ref{eq:ssm_iter_refine}.
\begin{align}
	O_{k} &= I_{k} \label{eq:ssm_init_feat} \\
	O_{k+mT} &= \psi_{\text{FA}}^{\left(L\right)}\left( \psi_{\text{FE}}^{\left(L\right)}(I_{k+mT}), \psi_{\text{FE}}^{\left(L\right)}(I_{k+(m-1)T})\right) \label{eq:ssm_iter_refine}
\end{align}
where the operators $\psi_{\text{FE}}^{(l)}$ and $\psi_{\text{FA}}^{(l)}$ denote the procedures of feature extraction and feature alignment, respectively, both with the same number of layers $L$. As illustrated in Fig.~\ref{fig:fbm} (b), during feature extraction, the input is processed by a downsample module at each layer. 

Each downsample module consists of two consecutive 2D convolutions with ReLU activation, where the strides are set to 2 and 1, respectively. Omitting the ReLU for brevity, the two convolutions with different strides are denoted as $\text{Conv}_{3\times3,s2}$ and $\text{Conv}_{3\times3,s1}$. The downsampling process is formally expressed in Eq.~\ref{eq:feature_extraction_downsample}. 
\begin{equation}
	\mathbf{F}_{\text{downsample}}(\mathbf{X}) = \text{Conv}_{3\times3,s2}(\text{Conv}_{3\times3,s1}(\mathbf{X}))
	\label{eq:feature_extraction_downsample}
\end{equation}

The feature maps of the neighbor at different scales $\left\{\mathbf{N}_1,\mathbf{N}_2,\dots,\mathbf{N}_L\right\}$ and those of the reference $\left\{\mathbf{R}_1,\mathbf{R}_2,\dots,\mathbf{R}_L\right\}$ are then jointly fed into the procedure of feature alignment, where the scale of feature maps decreases as the index increases.

As shown in Fig.~\ref{fig:fbm} (c), during the procedure of feature alignment, the feature maps from the reference input and the neighbor input are progressively aligned from the bottommost feature map up to the top. The entire process consists of $L$ steps. The core operation involved is deformable convolutions \cite{dai2017deformable, huang2019batching}, which more effectively capture variations in pose, local deformation, and scale of objects. Omitting ReLU and other activations, let $\mathcal{C}^{(n)}$ denote $n$ consecutive $3\times3$ convolutions with stride $1$, and let the deformable convolution operation be denoted by the operator $\theta\left(\cdot \right)$. At each step $l$, where $l\in\left[2,L\right]$, an intermediate output $\mathbf{S}_l$ is generated after upsampling at the end of the step. Finally, the last step generates the aligned feature output $\mathbf{O}_{\text{alignment}}$. 

The formulas for layer $L$ are shown in Eq.~\ref{eq:step_bottom_L} and Eq.~\ref{eq:step_up_L}.
\begin{align}
	\mathbf{B}_L&=\mathcal{C}^{\left(2\right)}\left(\text{Concat}\left(\mathbf{N}_L,\mathbf{R}_L\right)\right) \label{eq:step_bottom_L} \\
	\mathbf{S}_L&=\text{Up}\left(\mathcal{C}^{\left(1\right)}\left(\theta\left(\mathbf{B}_L,\mathbf{R}_L\right)\right)\right) \label{eq:step_up_L}
\end{align}

For layers $2$ to $L-1$, the model performs feature alignment by leveraging the upsampled output from the previous step. Let $l\in\left[2,L-1\right]$; the corresponding formulas are given in Eq.~\ref{eq:step_bottom_x} and Eq.~\ref{eq:step_up_x}.
\begin{align}
	\mathbf{B}_l&=\mathcal{C}^{\left(2\right)}\left(\text{Concat}\left(\mathcal{C}^{\left(1\right)}\left(\text{Concat}\left(\mathbf{N}_l,\mathbf{R}_l\right)\right),\mathbf{S}_{l+1}\right)\right) \label{eq:step_bottom_x} \\
	\mathbf{S}_l&=\text{Up}\left(\mathcal{C}^{\left(1\right)}\left(\text{Concat}\left(\theta\left(\mathbf{B}_l,\mathbf{R}_l\right),\mathbf{S}_{l+1}\right)\right)\right) \label{eq:step_up_x}
\end{align}

The final aligned feature output \(\mathbf{O}_{\text{alignment}}\) is obtained as shown in Eq.~\ref{eq:step_bottom_1} and Eq.~\ref{eq:step_up_1}.
\begin{align}
	&\mathbf{B}_1=\mathcal{C}^{\left(2\right)}\left(\text{Concat}\left(\mathcal{C}^{\left(1\right)}\left(\text{Concat}\left(\mathbf{N}_1,\mathbf{R}_1\right)\right),\mathbf{S}_{2}\right)\right) \label{eq:step_bottom_1} \\
	&\mathbf{O}_{\text{alignment}}=\mathcal{C}^{\left(1\right)}\left(\text{Concat}\left(\theta\left(\mathbf{B}_1,\mathbf{R}_1\right),\mathbf{S}_{2}\right)\right) \label{eq:step_up_1}
\end{align}

 After processing all $M_k + 1$ frames in $\mathcal{G}_k$, the group yields a set of refined output features: 
\begin{equation}
	\mathcal{O}_k= \left\{O_{k}, O_{k+T},\dots,O_{k+M_kT} \right\}
	\label{eq:output_group}
\end{equation}

\subsubsection{Temporal Reassembly}
In the final stage, the outputs from all $T$ groups are interleaved and sorted according to their original temporal indices to reconstruct a complete, coherent feature sequence of length $D$:
\begin{equation}
	\mathbf{Y}_{\text{SSM}} = \text{Merge}\bigl( \mathcal{O}_1, \mathcal{O}_2, \dots, \mathcal{O}_T \bigr)
	\label{eq:reassemble}
\end{equation}
where $\text{Merge}(\cdot)$ is a simple concatenation and sorting operation that reconstructs the full feature sequence $\{O_1, O_2, \dots, O_D\}$, preserving the temporal order.

\section{EXPERIMENT} \label{sec:experiment}
In this section, we first describe the experimental setup, including evaluation metrics and implementation details. Then, we evaluate our method against other baseline and state-of-the-art approaches on both SIRST and MIRST tasks. Finally, we present ablation studies to validate the effectiveness of our design.
\subsection{Datasets and Implementation}
\subsubsection{Datasets}To evaluate the performance of the proposed FeedbackSTS-Det, we employ two publicly available benchmark datasets. The NUDT-MIRSDT \cite{li2023direction} dataset comprises 120 sequences with 12,000 images, covering diverse infrared scenes such as clouds, oceans, and land surfaces. Following the original setup, each sequence is generated by applying local perturbations and synthetic noise to real infrared imagery. The IRSatVideo-LEO \cite{ying2025infrared} dataset contains 200 sequences with 91,022 images, constructed by combining real satellite imagery with simulated satellite motion trajectories. Both datasets incorporate synthetic augmentations, specifically perturbation injection in NUDT-MIRSDT and trajectory simulation in IRSatVideo-LEO, while providing high-quality pixel-level mask annotations. They have been widely adopted as benchmarks for multi-frame infrared small target detection. Importantly, to assess the generalization capability of our method, we conduct all evaluations strictly following the standard protocols without any dataset-specific customization. We further validate the robustness of our model under different sampling configurations and sequence lengths in the ablation studies (Sec. \ref{sec:ablation}).

\subsubsection{Metrics} In this paper, we adopt a mainstream evaluation framework \cite{zhang2023attention, li2022dense} for infrared small target detection models, incorporating pixel-level metrics (Intersection over Union (IoU), F-measure, False Alarm Rate (Fa)) and object-level metrics (Detection Rate (Pd), along with the Receiver Operating Characteristic (ROC)).
\subsubsection{Implementation Details}
For training, we use Soft-IoU loss \cite{huang2019batching} and the Adam optimizer with a MultiStepLR scheduler, starting at $1e^{-5}$ and halving at epochs $\left[5,10,15,20,25,30\right]$. Training lasts 30 epochs from scratch, implemented in PyTorch on six NVIDIA RTX 4090 GPUs. Additionally, both datasets then undergo the same augmentations, including random cropping, horizontal flipping, vertical flipping, and rotation, before being resized to $256 \times 256$. 

In terms of data preprocessing, both datasets then undergo the same augmentations (horizontal/vertical/channel flips, transpositions) before resizing to $256 \times 256$. Moreover, to ensure a consistent evaluation protocol, We exclude the “Mix” folder from NUDT-MIRSDT, which contains simulated sequences, to focus evaluation on authentic infrared sequences from its “images” and “masks” folders. To ensure a consistent evaluation protocol, We exclude the “Mix” folder from NUDT-MIRSDT, which contains simulated sequences, to focus evaluation on authentic infrared sequences from its “images” and “masks” folders.

Regarding the network architecture, we set the number of layers $L$ in both the feature extraction and feature alignment procedures of the BFBM, as mentioned in Section~\ref{sec:sparse_semantic_module}, to $2$, which achieves a good balance between alignment accuracy and computational cost. Furthermore, the sampling step $T$ of the SSM is a key hyperparameter. While larger values of $T$ can further reduce computational cost, excessively reducing the number of propagation steps inevitably compromises the effectiveness of feature propagation, leading to degraded detection performance. Therefore, $T$ must balance computational efficiency with sufficient temporal feature propagation. Our subsequent experiments demonstrate that values of 2, 3, or 4 yield optimal performance, effectively trading off between temporal coverage and processing efficiency.

\subsection{SOTA Comparisons}
We compare the proposed FeedbackSTS-Det against state-of-the-art methods in both single-frame and multi-frame infrared small target detection. For single-frame infrared small target detection, evaluated methods include three model-driven approaches: TopHat \cite{tom1993morphology}, NRAM \cite{zhang2018infrared}, and PSTNN \cite{zhang2019infrared}; and seven data-driven methods: ACM \cite{dai2021asymmetric}, ALCNet \cite{dai2021attentional}, DNANet \cite{li2022dense}, RDIAN \cite{sun2023receptive}, ISTDU-Net \cite{hou2022istdunet}, RPCANet \cite{wu2024rpcanet}, and L2SKNet$\_$FPN \cite{wu2024saliency}. For multi-frame infrared small target detection, comparisons include seven model-driven methods (MSLSTIPT \cite{sun2020infrared}, NFTDGSTV \cite{liu2023infrared}, RCTV \cite{liu2023representative}, IMNN-LWEC \cite{luo2022imnn}, SRSTT \cite{li2023sparse}, ASTTV-NTLA \cite{liu2021nonconvex}, 4DISDT \cite{wu2023infrared}) and five data-driven methods (DNANet$\_$DTUM \cite{li2023direction}, ALCNet$\_$DTUM \cite{li2023direction}, ResUNet$\_$DTUM \cite{li2023direction}, ACM$\_$RFR \cite{ying2025infrared}, ResUNet$\_$RFR \cite{ying2025infrared}). 

For data-driven MIRST methods, all metrics follow a uniform input sequence length of 10. Additionally, our networks with different sampling intervals $t$ are denoted as FeedbackSTS-Det-T$t$. When $t=1$, the model bypasses interval sampling in SSM. All models are retrained from scratch on both IRSatVideo-LEO \cite{ying2025infrared} and NUDT-MIRSDT \cite{li2023direction} datasets under identical settings for fair comparison.

\subsubsection{Quantitative Results}
Table \ref{tab:sota} presents a comprehensive comparison of state-of-the-art methods on the IRSatVideo-LEO \cite{ying2025infrared} and NUDT-MIRSDT \cite{li2023direction} using multiple evaluation metrics. Our proposed FeedbackSTS-Det variants consistently demonstrate superior performance across both datasets, with each sampling interval configuration exhibiting distinct advantages. The FeedbackSTS-Det-T2 variant achieves the most balanced and outstanding performance overall. On the NUDT-MIRSDT \cite{li2023direction} dataset, it obtains the best results in four key metrics: $mIoU$, $F_1$, $P_d$, and $AUC$, while maintaining the second-lowest false alarm rate. Similarly, on the IRSatVideo-LEO \cite{ying2025infrared} dataset, it delivers competitive detection accuracy with an exceptionally low false alarm rate of $5.5\times10^{-7}$, indicating its strong robustness in practical scenarios. The FeedbackSTS-Det-T3 and FeedbackSTS-Det-T4 variants also show remarkable capabilities, particularly in specific metrics. FeedbackSTS-Det-T3 achieves the second-best performance in $P_d$ and $AUC$ on NUDT-MIRSDT \cite{li2023direction}, while FeedbackSTS-Det-T4 excels on IRSatVideo-LEO \cite{ying2025infrared} with the highest $P_d$ of 0.9648 and $AUC$ of 0.9816. This performance progression across sampling intervals reveals an important trade-off: larger intervals reduce computational complexity while maintaining competitive detection accuracy.

Notably, all three FeedbackSTS-Det variants significantly outperform both traditional model-driven methods and contemporary data-driven approaches. While methods like DNANet$\_$DTUM \cite{li2023direction} and ResUNet$\_$RFR \cite{ying2025infrared} show respectable results, they are consistently surpassed by our proposed framework across virtually all metrics. This performance advantage is achieved while maintaining reasonable computational requirements, with all FeedbackSTS-Det variants utilizing only 5.68M parameters and FLOPs decreasing from 79.39G to 68.50G as the sampling interval increases. The consistent superiority of FeedbackSTS-Det across different sampling configurations validates the effectiveness of our architectural design and demonstrates its robustness for infrared small target detection in diverse operational scenarios.

\begin{figure}[!t]
	\centering
	\subfloat[]{%
		\includegraphics[width=0.51\linewidth]{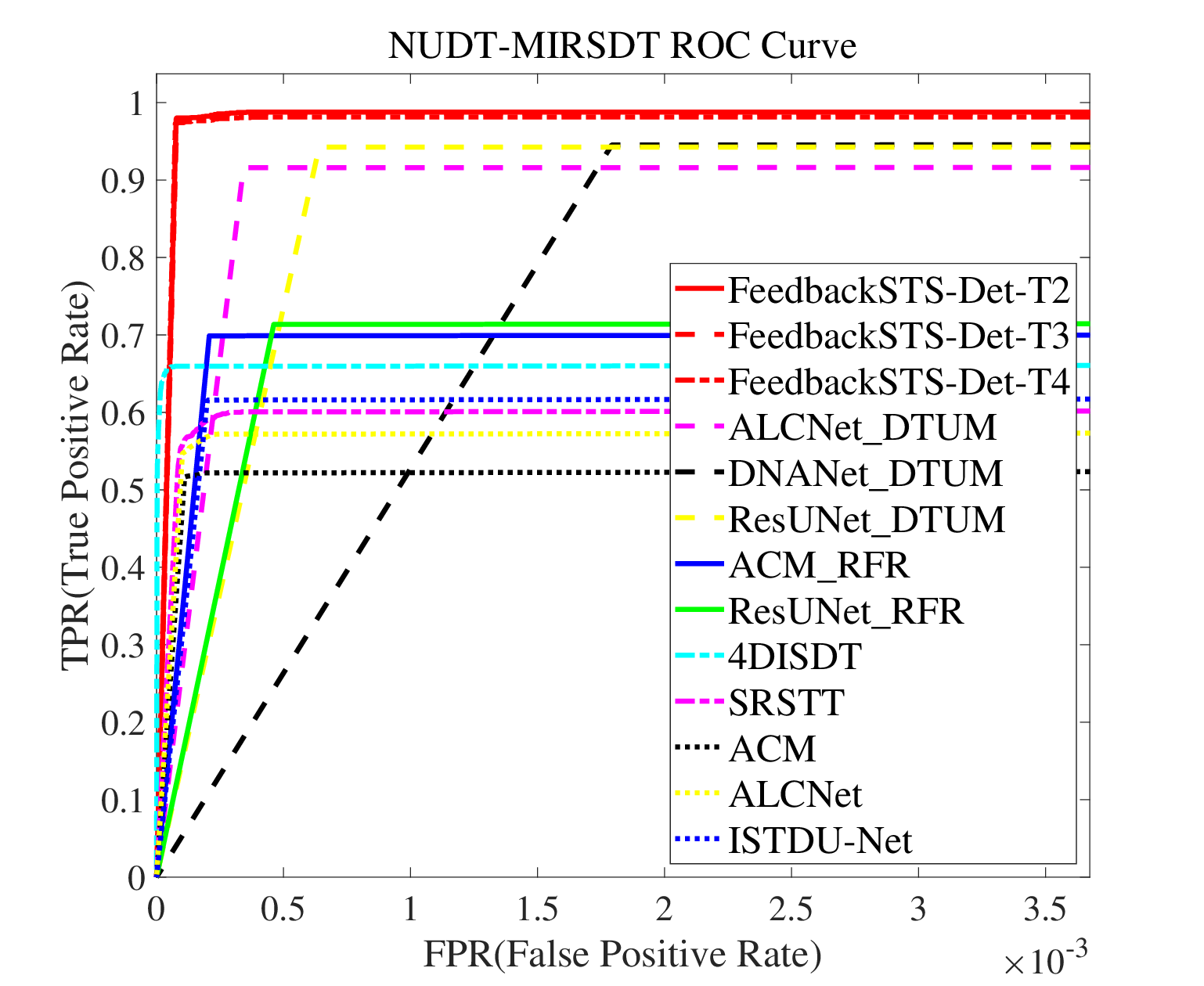}%
		\label{fig:ROC_NUDT_MIRSDT}%
	}%
	\subfloat[]{%
		\includegraphics[width=0.51\linewidth]{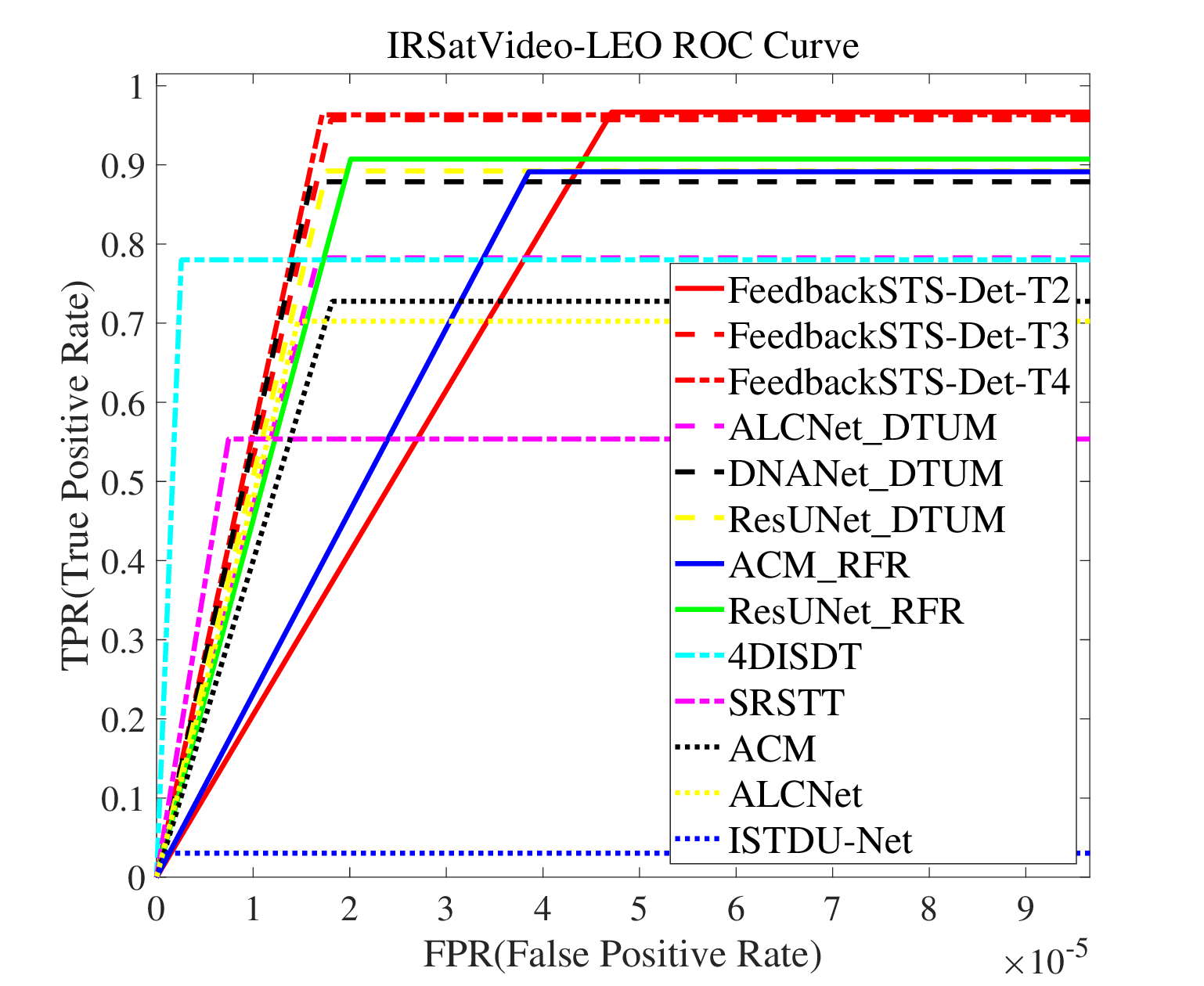}%
		\label{fig:ROC_IRSatVideo-LEO}%
	}%
	\caption{ROC curves on two benchmark datasets. (a) ROC Curve on NUDT-MIRSDT. (b) ROC Curve on IRSTVideo-LEO.}
	\label{fig:ROC}
\end{figure}

The ROC results are presented in Fig.~\ref{fig:ROC}. As shown in Fig. \ref{fig:ROC}(a), all FeedbackSTS-Det variants achieve a faster increase in TPR, reaching 1 more quickly than other models. In the more challenging IRSatVideo-LEO dataset (Fig. \ref{fig:ROC}(b)), although FeedbackSTS-Det-T2 exhibits a slower initial TPR rise, it ultimately reaches the maximum value faster. Meanwhile, FeedbackSTS-Det-T3 and FeedbackSTS-Det-T4 maintain strong TPR growth throughout the entire phase, confirming their robust detection performance.

\begin{table*}[!t]
\caption{The results of $mIoU\left(\times10^{-2}\right)$, $F_1\left(\times10^{-2}\right)$, $P_d\left(\times10^{-2}\right)$, $F_a\left(\times10^{-6}\right)$ and $AUC\left(\times10^{-2}\right)$ are achieved by different methods on the datasets IRSatvideo-LEO\cite{ying2025infrared} and NUDT-MIRSDT \cite{li2023direction}. "Params." represents the number of parameters. FLOPs are calculated based on an input sequence scale of $11\times256\times256$. The best-performing results are indicated in red font, while the second-best results are shown in blue font.}
    \label{tab:sota}
    \centering
    \renewcommand{\arraystretch}{1.5}
    \setlength{\tabcolsep}{3.5pt} 
    \scriptsize
    \begin{tabular}{c|cc|ccccc|ccccc}
        \hline
        \multirow{2}{*}{\centering \textbf{Method}} & \multirow{2}{*}{\centering \textbf{Params(M)}} & \multirow{2}{*}{\centering \textbf{FLOPs(G)}} & \multicolumn{5}{c|}{\textbf{NUDT-MIRSDT}} & \multicolumn{5}{c}{\textbf{IRSatVideo-LEO}} \\ \cline{4-13}
        ~ & ~ & ~ & $\bm{mIoU}\uparrow$ & $\bm{F_1}\uparrow$ & $\bm{P_d} \uparrow$ & $\bm{F_a}\downarrow$ & $\bm{AUC}\uparrow$ & $\bm{mIoU}\uparrow$ & $\bm{F_1}\uparrow$ & $\bm{P_d}\uparrow$ & $\bm{F_a}\downarrow$ & $\bm{AUC}\uparrow$ \\ \hline
        TopHat \cite{tom1993morphology} & - & - & 4.32 & 8.29 & 15.42 & 70.20 & 68.66 & 4.18 & 8.03 & 27.46 & 3.26 & 63.73 \\
        NRAM \cite{zhang2018infrared} & - & - & 2.98 & 5.78 & 8.68 & 26.80 & 59.89 & 1.51 & 2.97 & 9.45 & \textcolor{red}{0.20} & 54.68 \\
        PSTNN \cite{zhang2019infrared} & - & - & 13.03 & 23.06 & 21.62 & 84.92 & 64.64 & 3.50 & 6.76 & 46.29 & 61.31 & 73.14 \\ \hline
        ACM \cite{dai2021asymmetric} & 0.398 & 4.426 & 36.94 & 53.95 & 51.00 & 39.91 & 76.08 & 30.53 & 46.78 & 72.87 & 7.33 & 86.38 \\
        ALCNet \cite{dai2021attentional} & 0.427 & 4.157 & 32.26 & 48.79 & 54.65 & 65.38 & 78.58 & 31.36 & 47.75 & 70.27 & 5.21 & 85.12 \\
        DNA-Net \cite{li2022dense} & 4.697 & 71.306 & 27.51 & 43.15 & 60.00 & 211.94 & 80.04 & 0.13 & 0.27 & 0.92 & \textcolor{blue}{0.21} & 50.44 \\
        RDIAN \cite{sun2023receptive} & 0.217 & 40.902 & 19.33 & 32.40 & 61.06 & 509.90 & 80.74 & 0.10 & 0.19 & 6.73 & 2009.97 & 53.67 \\
        ISTDU-Net \cite{hou2022istdunet} & 2.752 & 87.388 & 33.45 & 50.13 & 60.88 & 111.24 & 80.78 & 0.47 & 0.93 & 3.09 & 0.96 & 51.51 \\
        RPCANet \cite{wu2024rpcanet} & 0.680 & 490.255 & 0.02 & 0.05 & 10.47 & 495831.30 & 73.74 & 0.00 & 0.00 & 0.65 & 499730.39 & 63.20 \\
        L2SKNet\_FPN \cite{wu2024saliency} & 1.071 & 66.085 & 21.12 & 34.88 & 43.06 & 92.03 & 71.85 & 1.09 & 2.15 & 47.67 & 362.75 & 73.23 \\
        \hline
        MSLSTIPT \cite{sun2020infrared} & - & - & 13.76 & 24.19 & 21.66 & 125.26 & 76.25 & 4.68 & 8.94 & 36.71 & 12.41 & 68.35 \\
        NFTDGSTV \cite{liu2023infrared} & - & - & 2.41 & 4.71 & 5.50 & 307.22 & 64.46 & 0.99 & 1.97 & 50.16 & 191.84 & 75.06 \\
        RCTV \cite{liu2023representative} & - & - & 0.34 & 0.68 & 0.18 & \textcolor{red}{1.40} & 52.55 & 8.74 & 16.07 & 39.31 & 0.93 & 69.65 \\
        IMNN-LWEC \cite{luo2022imnn} & - & - & 6.04 & 11.40 & 13.51 & 209.38 & 64.57 & 1.51 & 2.97 & 46.00 & 265.21 & 72.97 \\
        SRSTT \cite{li2023sparse} & - & - & 12.18 & 21.72 & 21.16 & 444.36 & 80.01 & 12.68 & 22.51 & 55.40 & 6.63 & 77.68 \\
        ASTTV-NTLA \cite{liu2021nonconvex} & - & - & 0.78 & 1.55 & 0.36 & 52.54 & 53.84 & 0.80 & 1.59 & 50.88 & 206.00 & 75.42 \\
        4DISDT \cite{wu2023infrared} & - & - & 10.66 & 19.26 & 15.92 & 105.03 & 82.97 & 23.13 & 37.58 & 78.16 & 1.81 & 89.00 \\ \hline
        DNANet\_DTUM \cite{li2023direction} & 1.205 & 97.757 & 10.44 & 18.91 & 94.29 & 1726.77 & 97.18 & 38.88 & 55.99 & 88.24 & 6.31 & 93.93 \\
        ALCNet\_DTUM \cite{li2023direction} & 0.842 & 16.897 & 35.67 & 52.58 & 91.82 & 175.71 & 95.78 & 35.49 & 52.39 & 79.18 & 2.53 & 89.13 \\
        ResUNet\_DTUM \cite{li2023direction} & 0.298 & 22.772 & 24.33 & 39.13 & 95.65 & 499.13 & 97.09 & 38.22 & 55.30 & 89.52 & 6.58 & 94.61 \\
        ACM\_RFR \cite{ying2025infrared} & 0.504 & 40.842 & 29.98 & 46.13 & 69.24 & 135.79 & 84.92 & 24.68 & 39.59 & 89.18 & 16.88 & 94.56 \\
        ResUNet\_RFR \cite{ying2025infrared} & 1.012 & 79.801 & 30.98 & 47.31 & 70.94 & 286.67 & 85.64 & 37.49 & 54.54 & 90.82 & 6.02 & 95.36 \\ \hline
        FeedbackSTS-Det-T2 (Ours) & 5.678 & 79.386 & \textcolor{red}{52.24} & \textcolor{red}{68.63} & \textcolor{red}{97.41} & \textcolor{blue}{14.42} & \textcolor{red}{99.38} & \textcolor{red}{43.50} & \textcolor{red}{60.63} & 95.28 & 0.55 & 97.46 \\
        FeedbackSTS-Det-T3 (Ours) & 5.678 & 73.941 & 51.55 & 68.03 & \textcolor{blue}{97.29} & 19.23 & \textcolor{blue}{99.32} & 41.96 & 59.11 & \textcolor{blue}{95.83}  & 4.51 & \textcolor{blue}{97.84} \\
        FeedbackSTS-Det-T4 (Ours) & 5.678 & 68.496 & \textcolor{blue}{51.95} & \textcolor{blue}{68.38} & 96.65 & 18.45 & 99.05 & \textcolor{blue}{43.42} & \textcolor{blue}{60.54} & \textcolor{red}{96.48} & 3.07 & \textcolor{red}{98.16} \\ \hline
    \end{tabular}
\end{table*}

\subsubsection{Qualitative Results}
\begin{figure*}[!t]
	\centering
	\includegraphics[width=\textwidth]{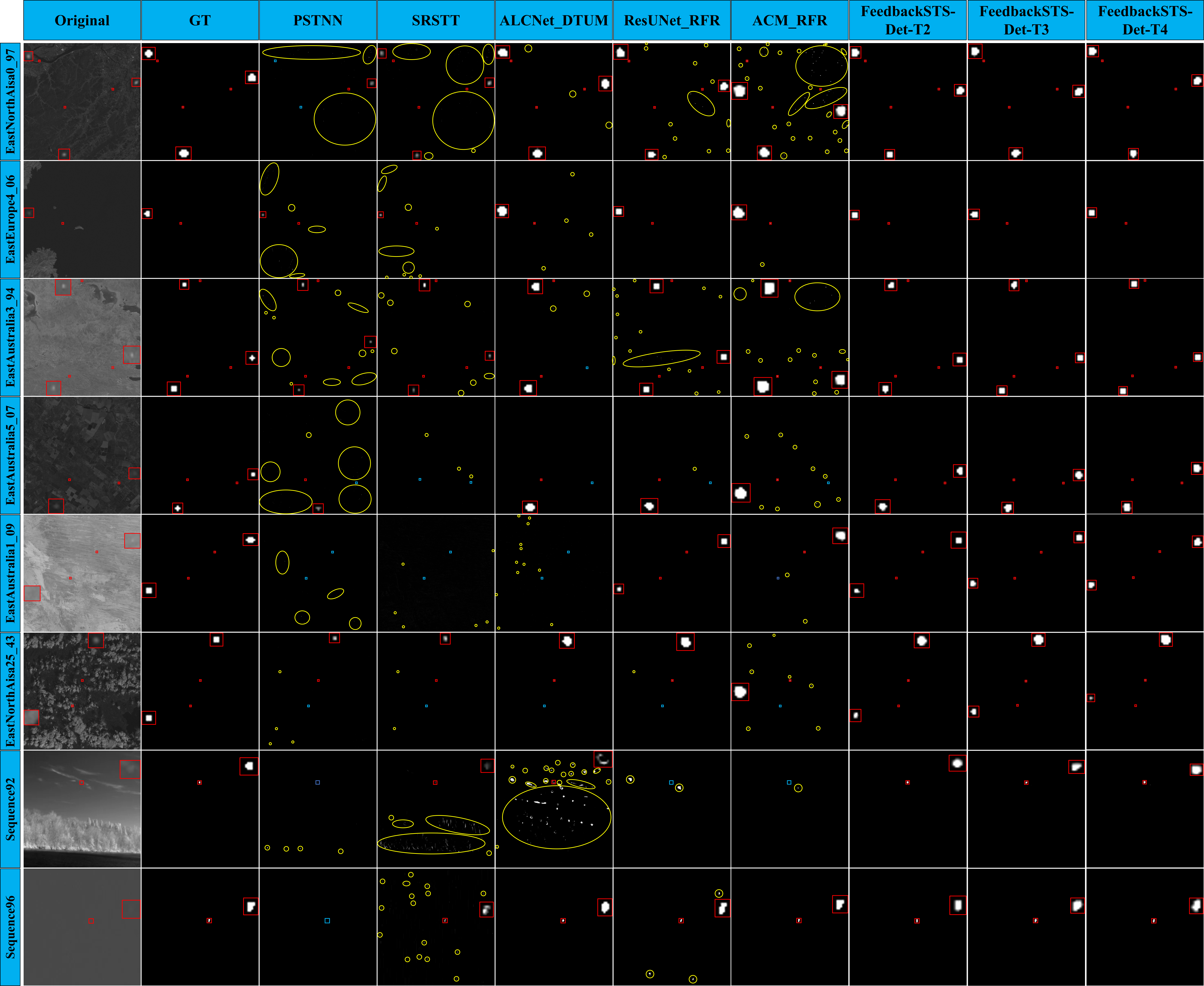}
	\caption{Qualitative results of different methods. For better visual presentation, the target regions are highlighted with red boxes and then displayed as zoomed-in views. Missed detections and false alarms are marked with blue and yellow circles, respectively.}
	\label{fig:Saliency}
\end{figure*}

Qualitative results are presented in Fig. \ref{fig:Saliency}. The selected examples are organized as follows: Rows 1--2 illustrate easy scenes from IRSatVideo-LEO \cite{ying2025infrared}; Rows 3--4 depict medium scenes; Rows 5--6 present challenging scenes from the same dataset; and Rows 7--8 show sequence samples from NUDT-MIRSDT \cite{li2023direction}. The scenes encompass satellite, aerial, and maritime imagery captured from a vertical perspective. It can be observed that traditional model-driven methods, including low-rank and sparse representation techniques such as PSTNN \cite{zhang2019infrared} and SRSTT \cite{li2023sparse}, exhibit limited detection performance in these scenes. The typical single-frame low-rank and sparse infrared small target detection method, PSTNN \cite{zhang2019infrared}, produces significant false alarms in most satellite nadir-view scenes except for the sequence "EastNorthAsia25$\_$43", and fails to detect targets correctly in "Sequence92" and "Sequence96". SRSTT \cite{li2023sparse}, as a representative multi-frame low-rank and sparse detection method, generates substantial false alarms across all sequences and exhibits missed detections in several satellite view sequences. Data-driven multi-frame methods show noticeable improvement, yet still exhibit certain limitations. Except for "Sequence96", ALCNet$\_$DTUM \cite{li2023direction} produces false alarms in all other sequences, with particularly severe cases in "Sequence92". RFR-based \cite{ying2025infrared} frameworks (e.g., ResUNet$\_$RFR \cite{ying2025infrared} and ACM$\_$RFR \cite{ying2025infrared}) yield significant false alarms in sequences such as "EastNorthAsia0$\_$97", "EastAustralia3$\_$94", "EastAustralia5$\_$07", and "EastNorthAsia25$\_$43". However, ACM$\_$RFR \cite{ying2025infrared} achieves relatively good detection performance on "Sequence96". In contrast, it can be observed that all of the variants of our FeedbackSTS-Det achieve the optimal performance across all scenarios and all difficulty levels of datasets, specifically in terms of accurate target localization, low false alarm rate, and the morphological accuracy of detected targets.

\subsection{Ablation Study} \label{sec:ablation}
This subsection presents experiments on FeedbackSTS-Det and its variants, aiming to verify the potential advantages of our proposed modules and key design decisions. If not stated explicitly, we set the temporal sampling interval to 2.
\subsubsection{Spatio-Temporal Semantic Feedback Strategy}
Introduced in Section \ref{sec:semantic_feedback_framework}, the spatio-temporal semantic feedback strategy is designed to strengthen inter-frame feature linkages for better infrared small target detection. Our approach is built upon a 3D U-Net backbone \cite{cciccek20163d}. A systematic comparison of the ablated variants of this feedback design is provided in Table \ref{tab:ablation_feature_iterative_configuration}. This ablation study is designed to validate the effectiveness of our proposed framework by contrasting it against several alternative configurations, including unidirectional (All-Fwd, All-Bwd), partial semantic feedback (Part-FB1, Part-FB2), and feedback-disabled schemes (Dec-NoFB, Enc-NoFB).

As summarized in Table \ref{tab:ablation_feature_iterative_configuration}, while simpler configurations like Enc-NoFB achieve the lowest parameter count and FLOPs, our optimal method, Full-FB, strikes a critical balance. It maintains full feature interaction capabilities in both the encoder and decoder, incurring only a modest increase in computational cost compared to the most constrained variants. Subsequent experiments confirm that the proposed spatio-temporal semantic feedback framework is a necessary and efficient component for accurate target localization, shape delineation, and false alarm suppression.

\begin{table*}[htbp]
	\centering
	\normalsize
	\caption{Ablation study on feedback design variants: configuration and efficienc. "Params." represents the number of parameters. FLOPs are calculated based on an input sequence scale of $5\times256\times256$. The row corresponding to the proposed spatio-temporal semantic feedback framework is highlighted with a gray background.}
	\label{tab:ablation_feature_iterative_configuration}
	\begin{tabular}{l l c c c c}
		\toprule
		\textbf{Function \& Characteristic} & \textbf{Abbr.} & \textbf{Encoder} & \textbf{Decoder} & \textbf{Params$\downarrow$} & \textbf{FLOPs$\downarrow$} \\
		\midrule
		Decoder without semantic feedback & Dec-NoFB & $\left[+,+,+,+,+\right]$ & $\left[\sim,\sim,\sim,\sim\right]$ & 5.046M & 26.979G\\
		Encoder without semantic feedback & Enc-NoFB & $\left[\sim,\sim,\sim,\sim,\sim\right]$ & $\left[-,-,-,-\right]$  & 2.748M & 22.684G\\
		All forward semantic feedback & All-Fwd & $\left[+,+,+,+,+\right]$ & $\left[+,+,+,+\right]$  & 5.678M & 30.144G\\
		All backward semantic feedback & All-Bwd & $\left[-,-,-,-,-\right]$ & $\left[-,-,-,-\right]$  & 5.678M & 30.144G\\
		Partial semantic feedback (Pattern 1) & Part-FB1 & $\left[+,\sim,+,\sim,+\right]$ & $\left[-,\sim,-,\sim\right]$  & 4.560M & 26.559G\\
		Partial semantic feedback (Pattern 2) & Part-FB2 & $\left[+,\sim,+,\sim,+\right]$ & $\left[\sim,-,\sim,-\right]$  & 4,876M & 25.343G\\
		\rowcolor{gray!20}
	    Full semantic feedback (Ours) & Full-FB & $\left[+,+,+,+,+\right]$ & $\left[-,-,-,-\right]$ & 5.678M & 30.144G \\
		\midrule
		\multicolumn{6}{l}{\quad\textbf{Symbols:} ~~$+$: FSTSRM~~~~~~$\sim$: 3D Conv Block~~~~~~$-$: BSTSRM} \\
		\bottomrule
	\end{tabular}
\end{table*}

For consistent performance evaluation where higher values indicate better results across all metrics, including $mIoU$, $F_1$, $P_d$, and $AUC$, we introduce the false alarm suppression rate (FSR), defined as follows:

\begin{equation}
	\label{equation:fsr}
	FSR=1-F_a\times10^3
\end{equation}

\begin{figure*}[htbp]
	\centering
	\setlength{\tabcolsep}{0.5pt} 
	\begin{tabular}{@{}ccc@{}}
		\subfloat[]{%
			\includegraphics[width=0.33\textwidth]{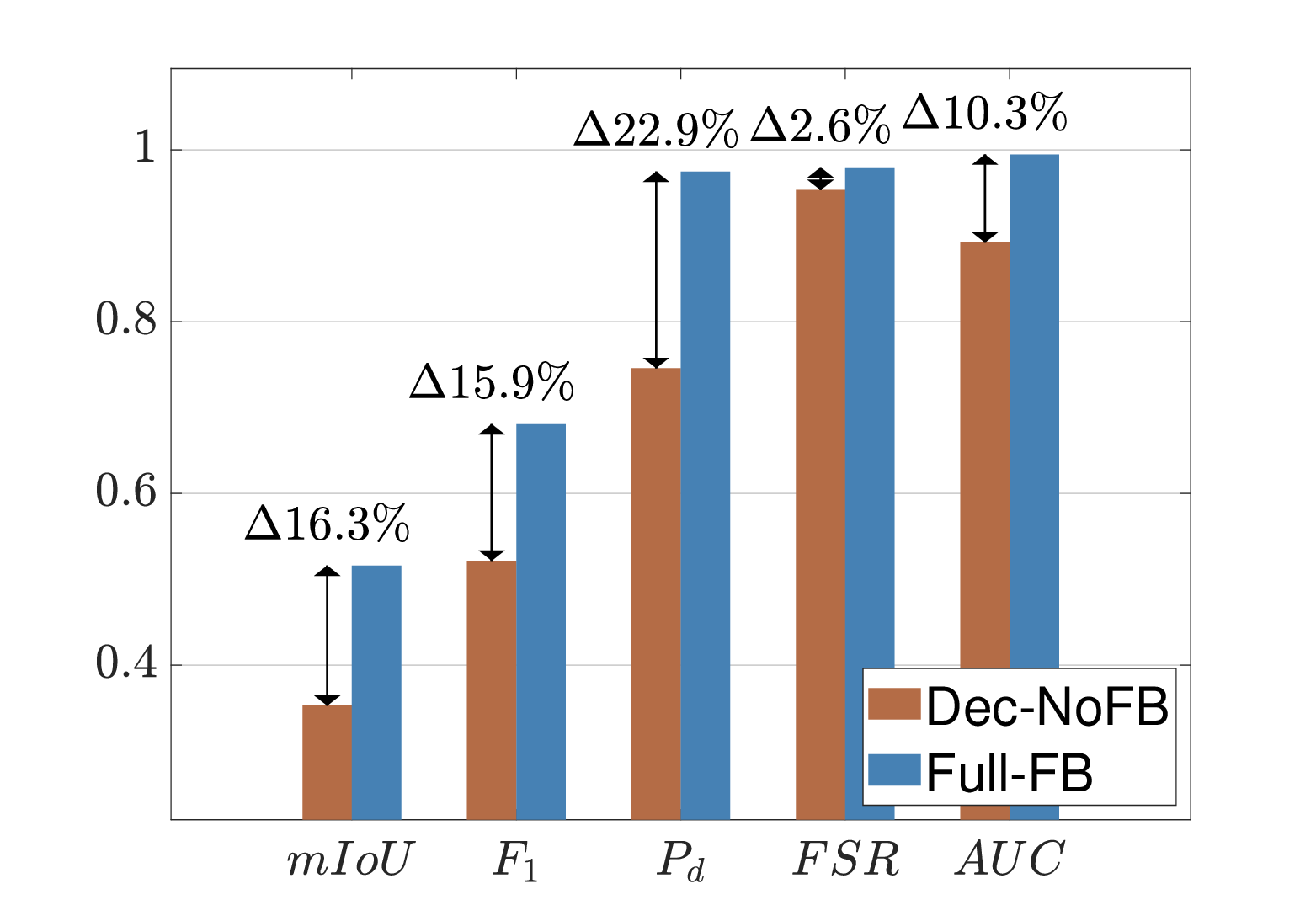}%
			\label{fig:feat_prop_sub1}%
		} &
		\subfloat[]{%
			\includegraphics[width=0.33\textwidth]{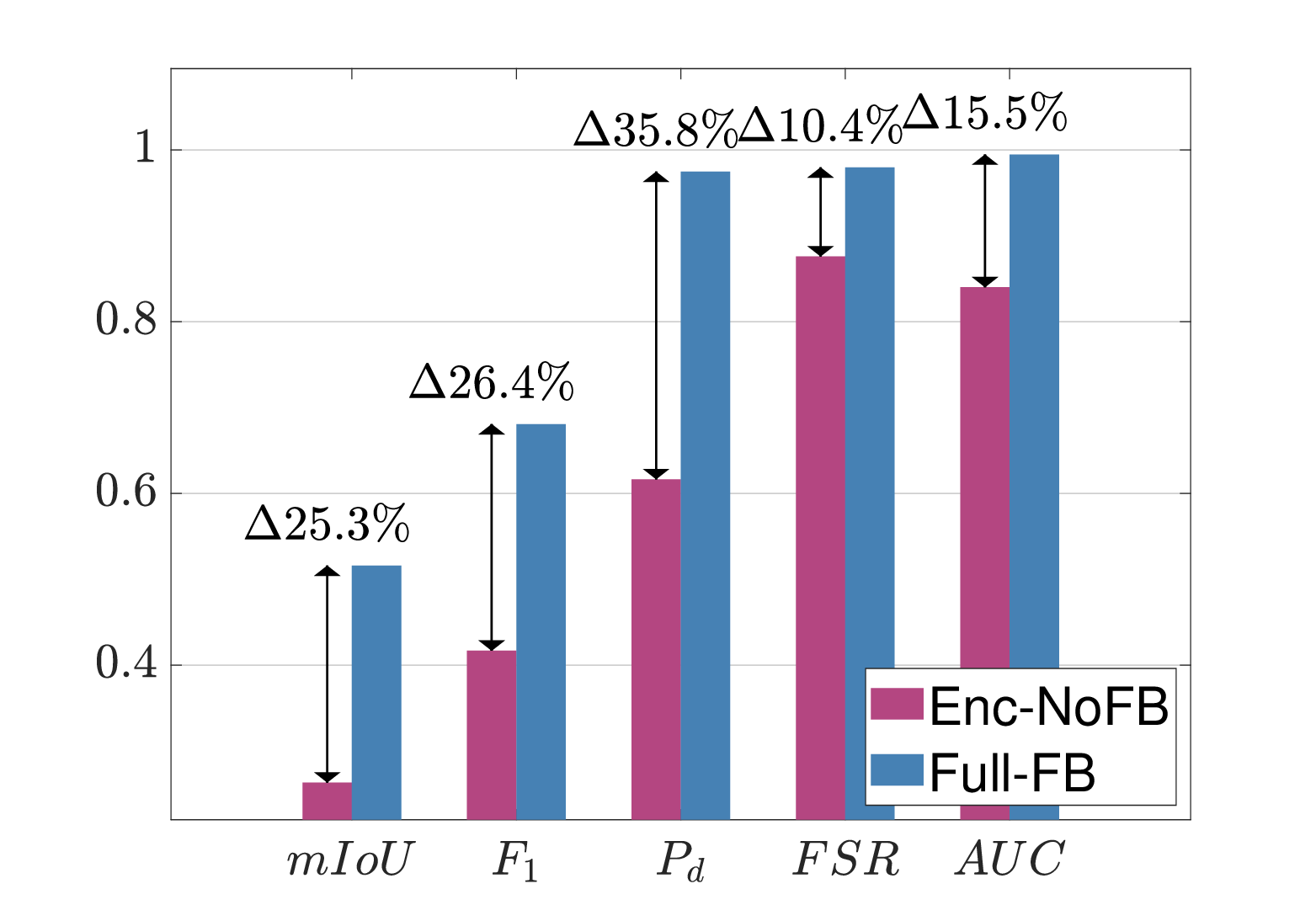}%
			\label{fig:feat_prop_sub2}%
		} &
		\subfloat[]{%
			\includegraphics[width=0.33\textwidth]{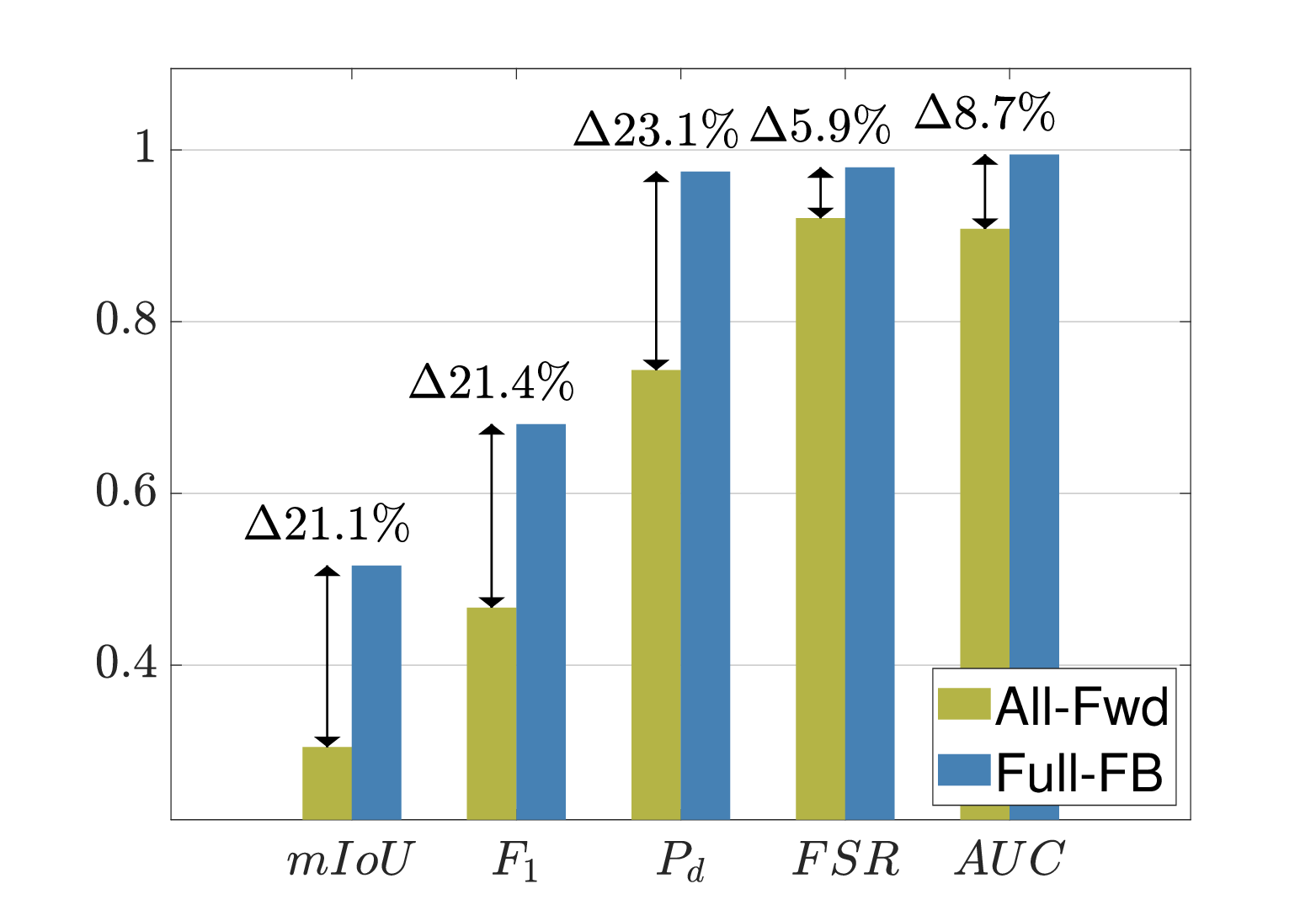}%
			\label{fig:feat_prop_sub3}%
		} \\[-6pt]
		\subfloat[]{%
			\includegraphics[width=0.33\textwidth]{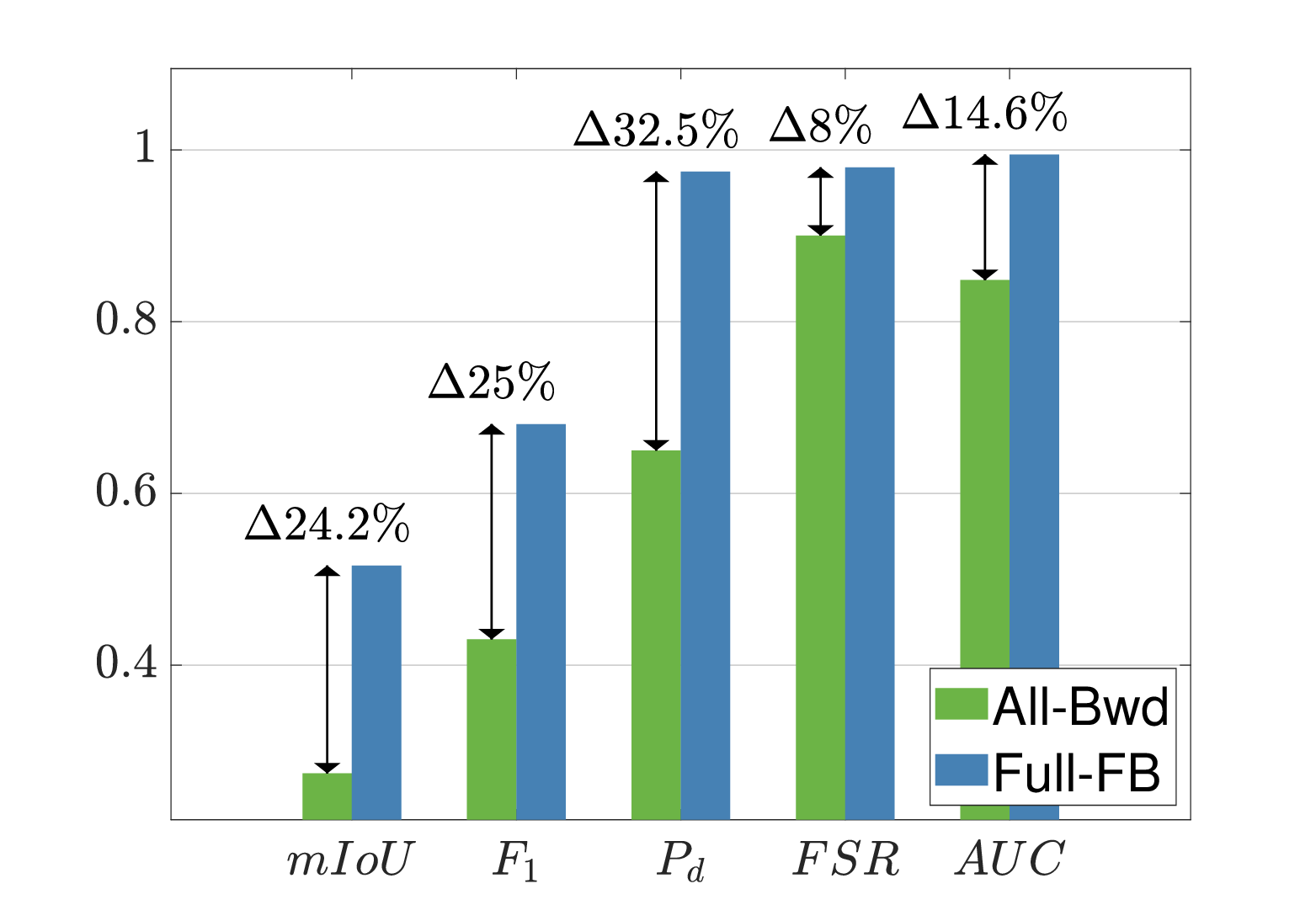}%
			\label{fig:feat_prop_sub4}%
		} &
		\subfloat[]{%
			\includegraphics[width=0.33\textwidth]{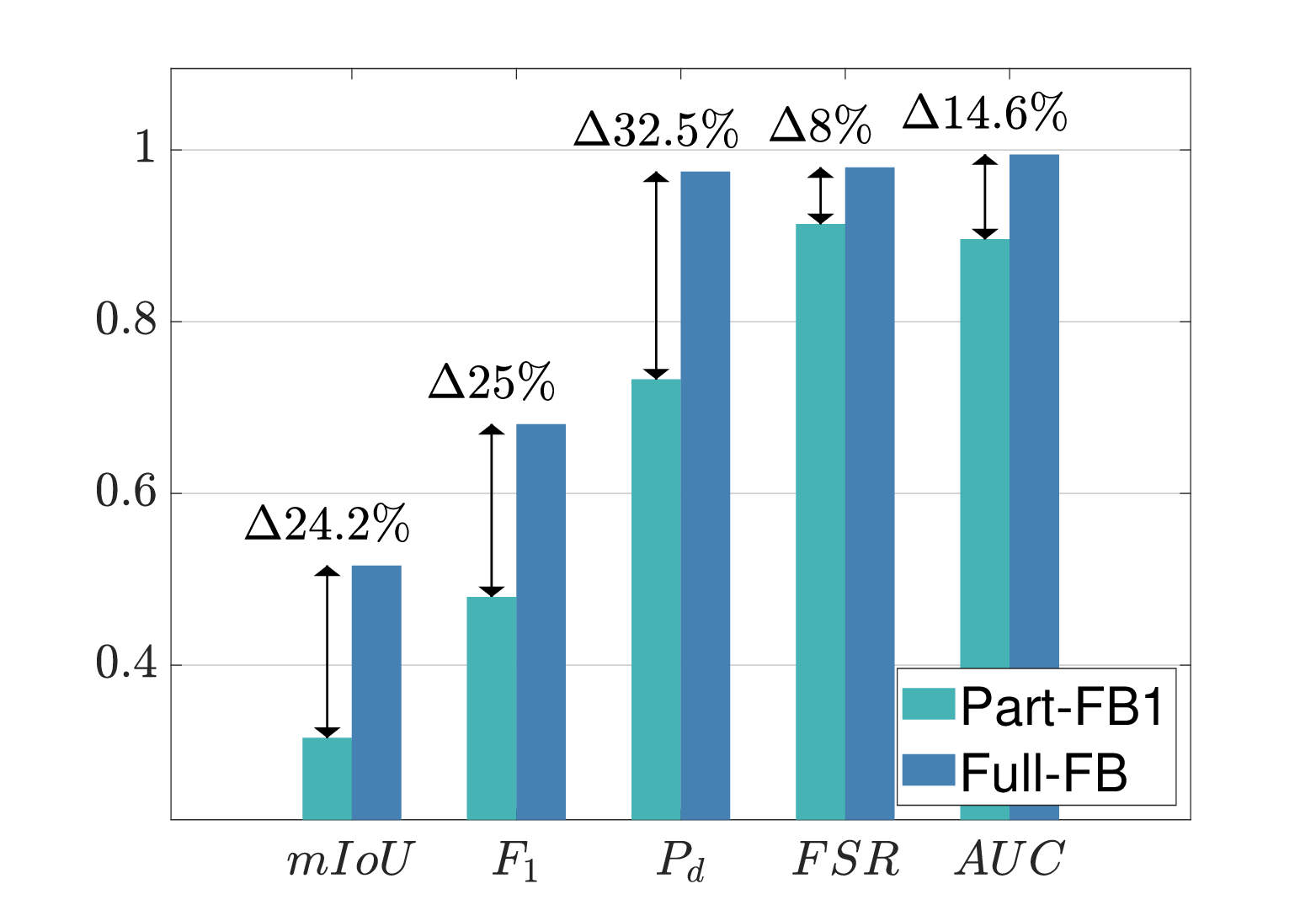}%
			\label{fig:feat_prop_sub5}%
		} &
		\subfloat[]{%
			\includegraphics[width=0.33\textwidth]{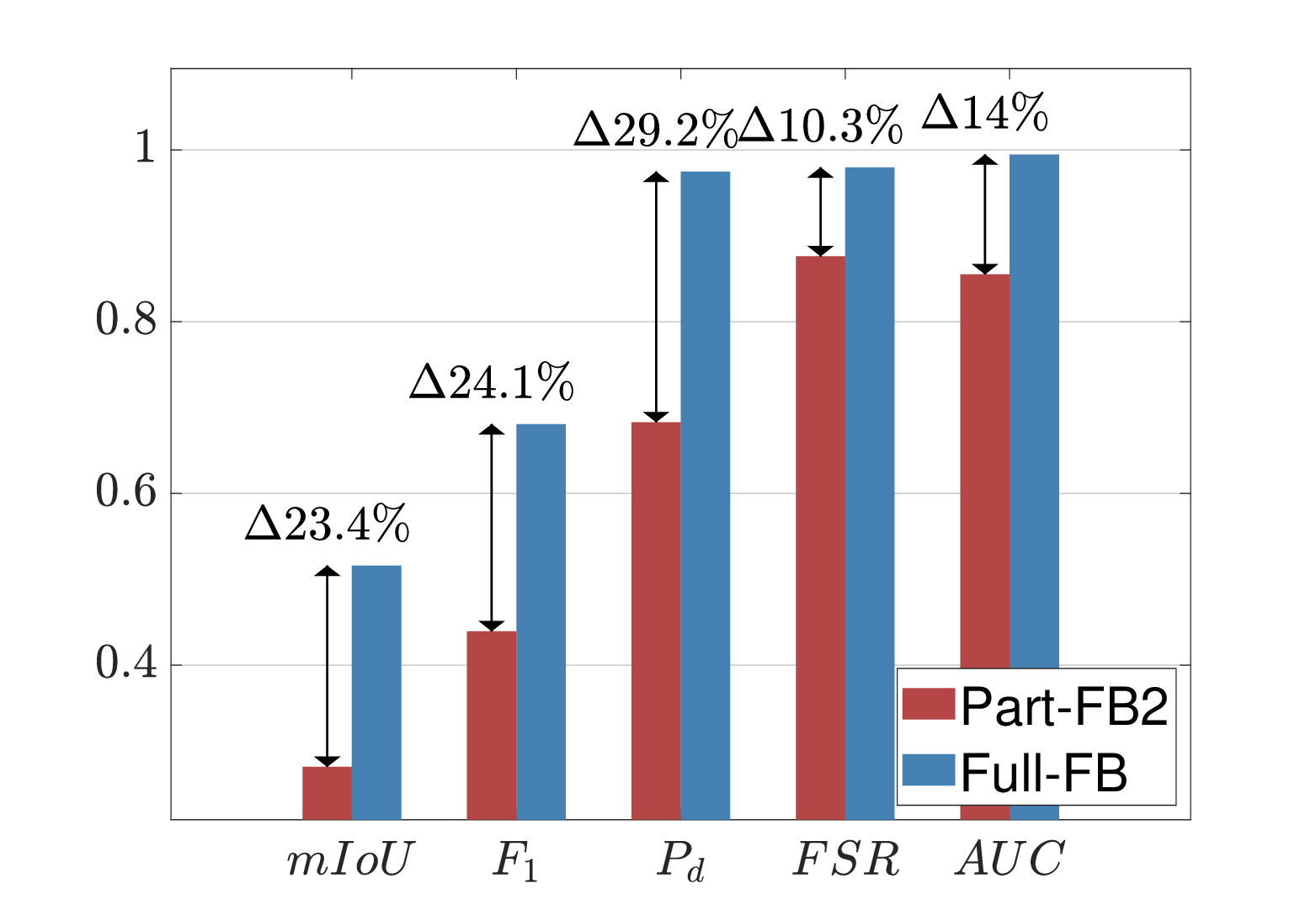}%
			\label{fig:feat_prop_sub6}%
		}
	\end{tabular}
	\vspace{-3pt}
	\caption{Ablation study of different feedback design variants against the Full-FB.  Evaluation metrics include mean Intersection over $mIoU$, $F_1$, $P_d$, $FSR$, and $AUC$, where $\Delta$ represents the performance difference relative to the Full-FB.(a) Dec-NoFB vs Full-FB. (b) Enc-NoFB vs Full-FB. (c) All-Fwd vs Full-FB. (d) All-Bwd vs Full-FB. (e) Part-FB1 vs Full-FB. (f) Part-FB2 vs Full-FB.}
	\label{fig:feat_prop}
\end{figure*}

As illustrated in Fig.~\ref{fig:feat_prop}, we benchmark all spatio-temporal semantic feedback variants from Table \ref{tab:ablation_feature_iterative_configuration} against our optimal method, Full-FB, on the NUDT-MIRSDT \cite{li2023direction} dataset. The evaluation uses five metrics ($mIoU$, $F_1$, $P_d$, $FSR$, $AUC$) under a fixed input size of $5\times256\times256$, and results show the clear superiority of our method across all metrics.

\begin{figure*}[htbp]
	\centering
	\setlength{\tabcolsep}{0.5pt} 
	\begin{tabular}{@{}cc@{}}
		\subfloat[]{%
			\includegraphics[width=0.44\textwidth]{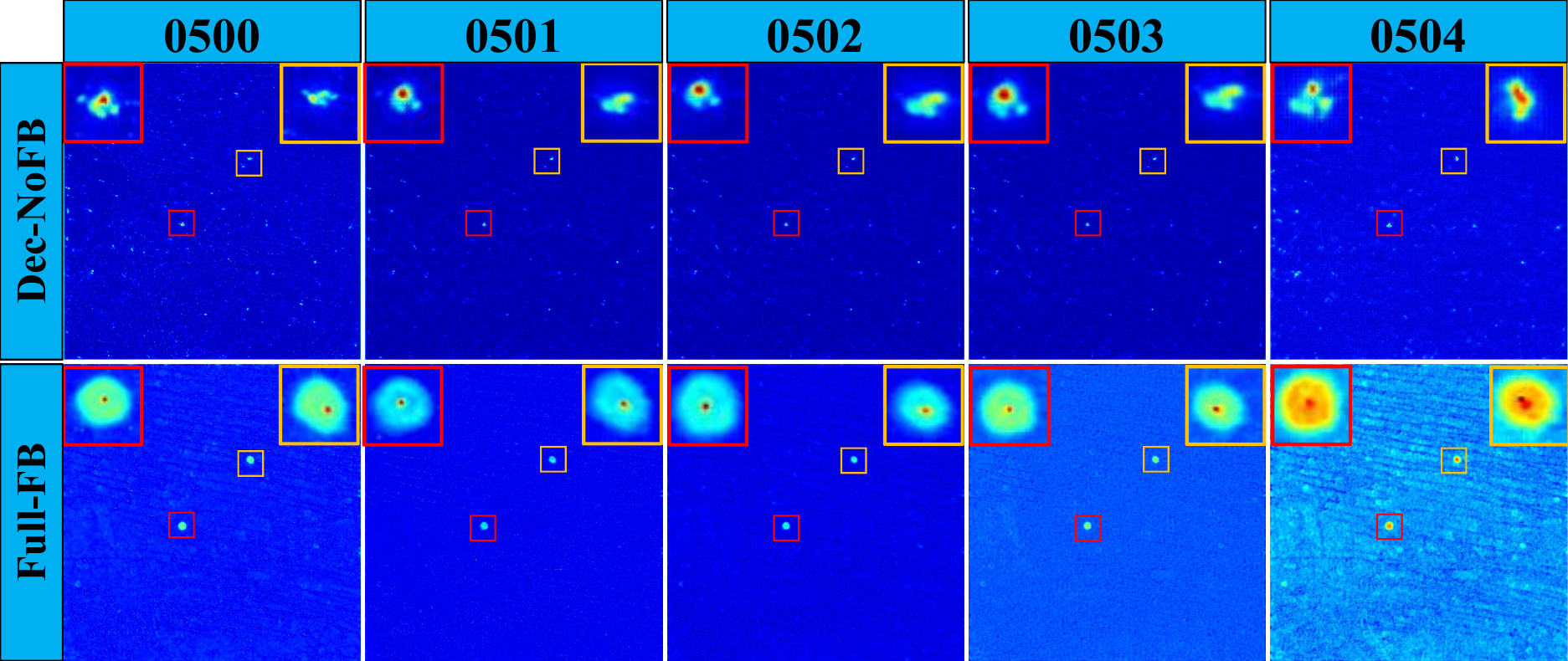}%
			\label{fig:feat_prop_color_sub1}%
		} &
		\subfloat[]{%
			\includegraphics[width=0.44\textwidth]{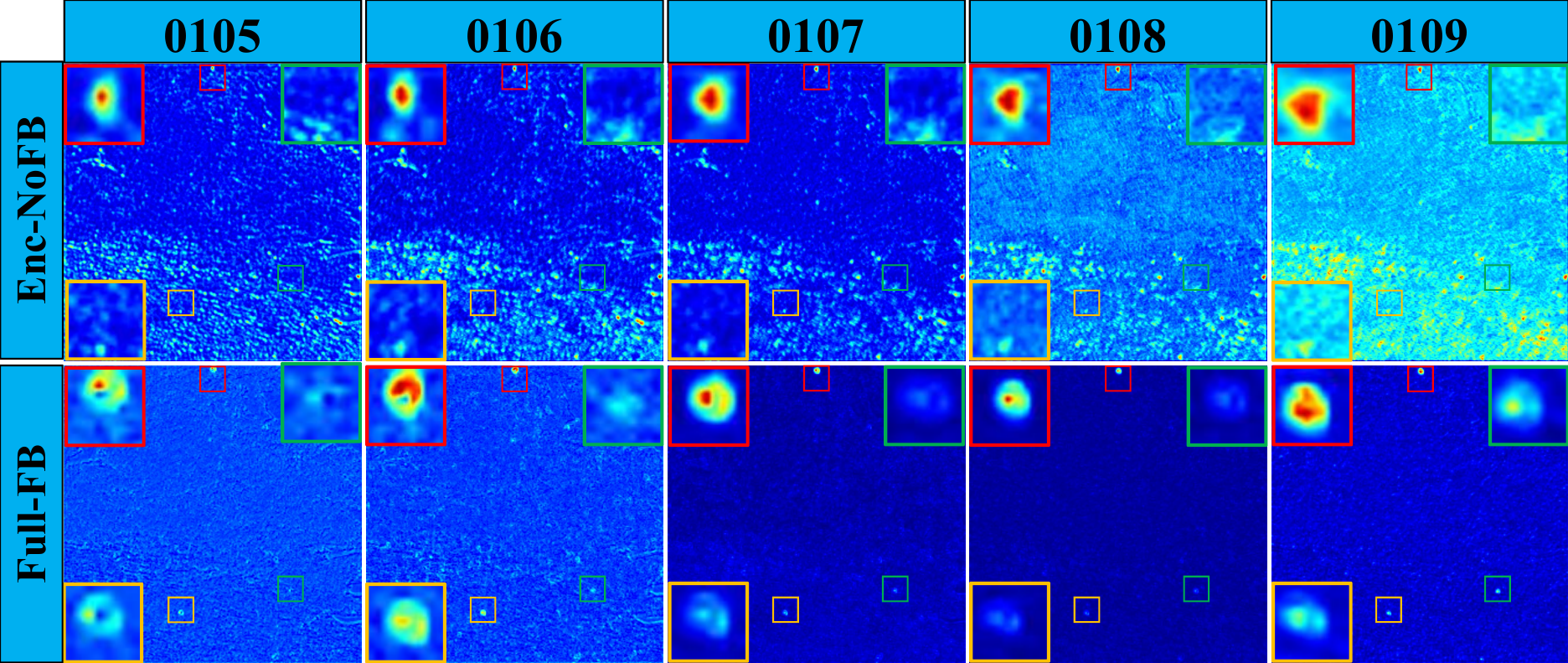}%
			\label{fig:feat_prop_color_sub2}%
		} \\[-6pt]
		\subfloat[]{%
			\includegraphics[width=0.44\textwidth]{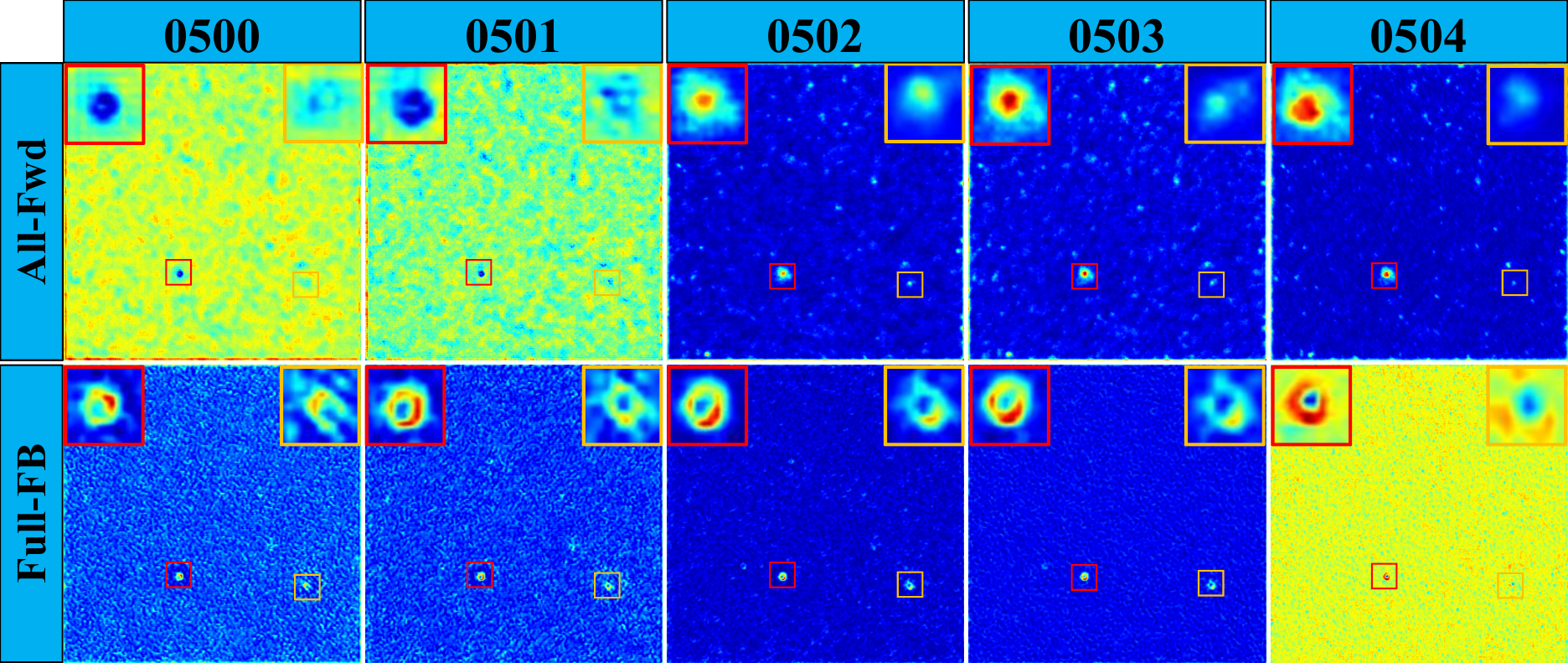}%
			\label{fig:feat_prop_color_sub3}%
		} &
		\subfloat[]{%
			\includegraphics[width=0.44\textwidth]{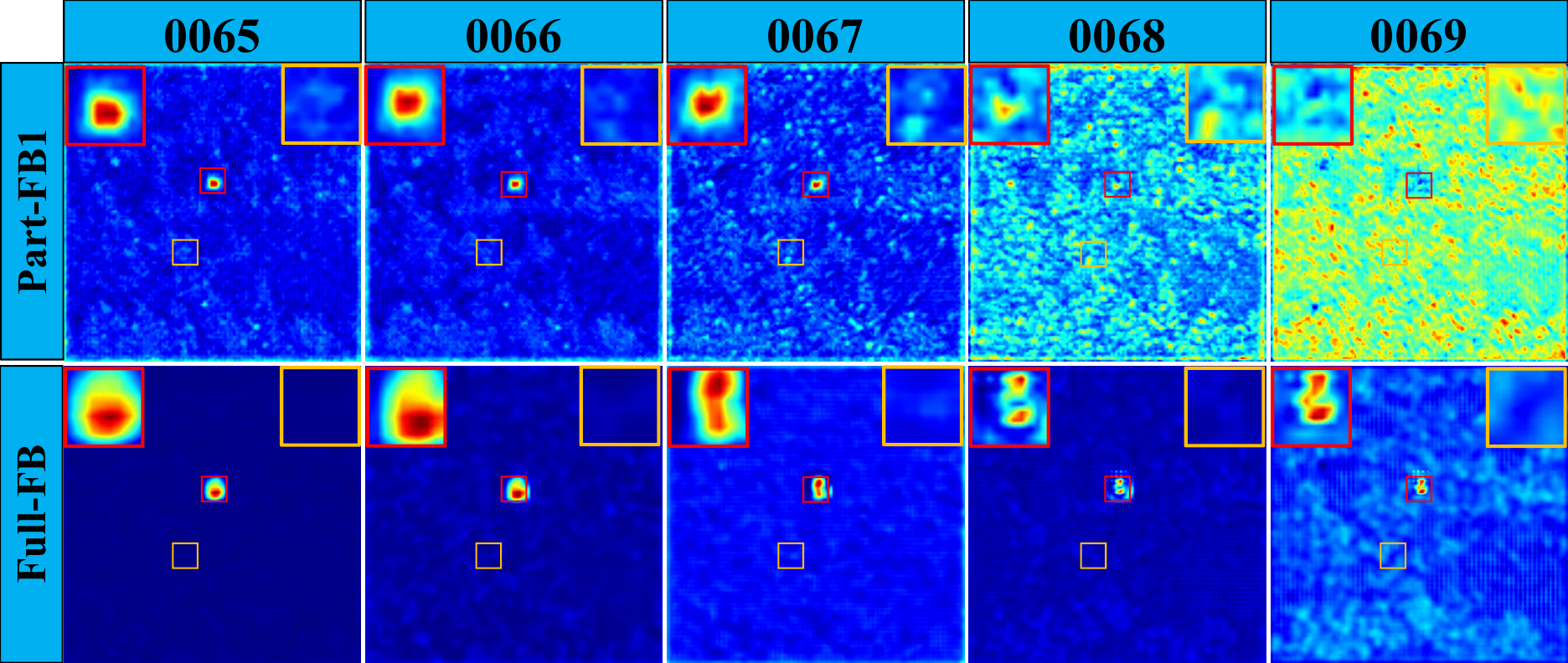}%
			\label{fig:feat_prop_color_sub4}%
		} \\[-6pt]
		\subfloat[]{%
			\includegraphics[width=0.44\textwidth]{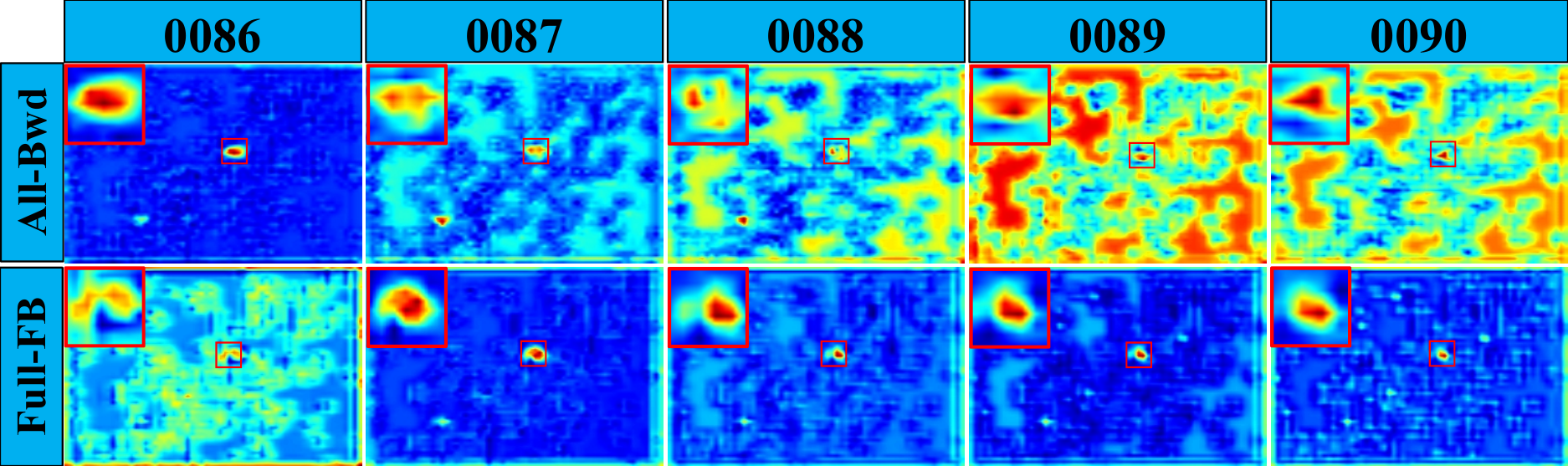}%
			\label{fig:feat_prop_color_sub5}%
		} &
		\subfloat[]{%
			\includegraphics[width=0.44\textwidth]{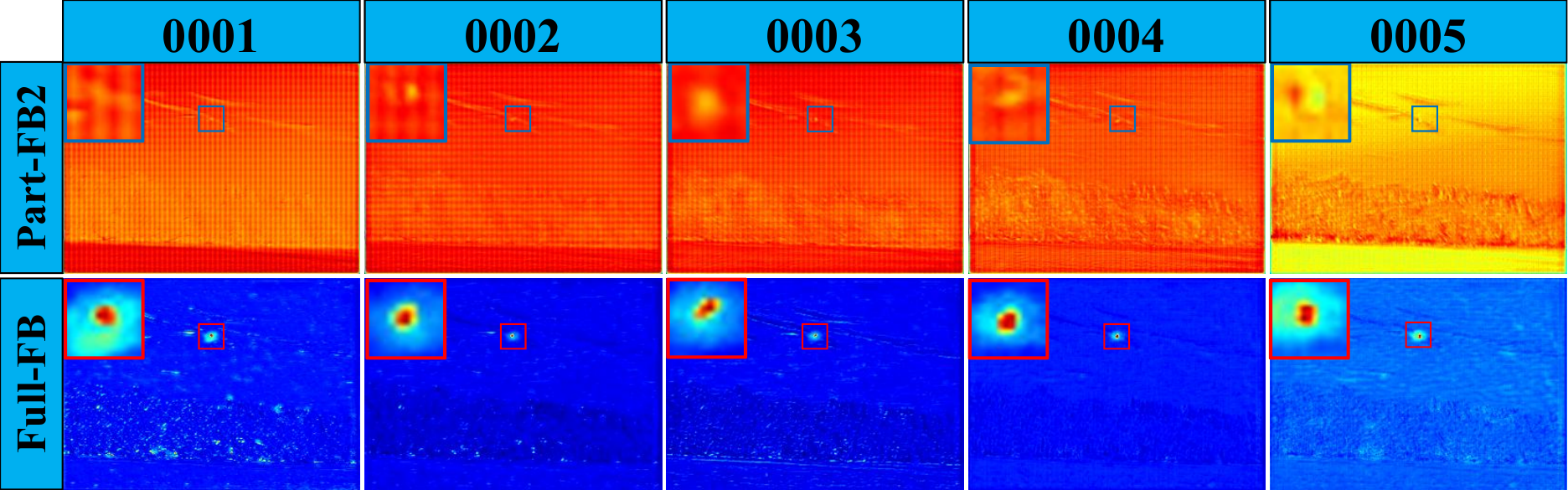}%
			\label{fig:feat_prop_color_sub6}%
		}
	\end{tabular}
	\vspace{-3pt}
	\caption{Ablation experiments show the visual comparison of different spatio-temporal semantic feedback approaches. (a) Visual comparison of Dec-NoFB versus Full-FB, evaluated on the IRSatVideo-LEO \cite{ying2025infrared}. (b) Visual comparison of Enc-NoFB versus Full-FB, evaluated on the IRSatVideo-LEO \cite{ying2025infrared}. (c) Visual comparison of All-Fwd versus Full-FB, evaluated on the IRSatVideo-LEO \cite{ying2025infrared}. (d) Visual comparison of Part-FB1 versus Full-FB, evaluated on the IRSatVideo-LEO \cite{ying2025infrared}. (e) Visual comparison of All-Bwd versus Full-FB, evaluated on the NUDT-MIRSDT \cite{li2023direction}. (e) Visual comparison of Part-FB2 versus Full-FB, evaluated on the NUDT-MIRSDT \cite{li2023direction}.}
	\label{fig:feat_prop_color}
\end{figure*}

The visual comparisons corresponding to the methods in Table~\ref{tab:ablation_feature_iterative_configuration} are provided in Fig.~\ref{fig:feat_prop_color}. For clarity, targets are annotated with bounding boxes positioned at image corners, with distinct colors assigned to different targets for easy cross‑reference. The effectiveness of backward feedback in the decoder is demonstrated by comparing Full‑FB with alternative designs. First, in Fig.~\ref{fig:feat_prop_color} (a), Full‑FB shows progressively stronger suppression and clearer targets during backward feedback (frames $\#0504$ to $\#0500$), while Dec‑NoFB performs poorly. Second, Fig.~\ref{fig:feat_prop_color} (b) reveals that the All‑Fwd method yields unclear targets in initial frames ($\#0500$, $\#0501$) during forward feedback, whereas Full‑FB exhibits this limitation only in the initial frame ($\#0504$) of its backward process. Moreover, in Fig.~\ref{fig:feat_prop_color} (c) and Fig.~\ref{fig:feat_prop_color}(e), targets in the second row (Full‑FB) are consistently clearer than those in the first row (the compared methods), highlighting the benefit of forward feedback in the encoder. Finally, Fig.~\ref{fig:feat_prop_color} (d) and Fig.~\ref{fig:feat_prop_color} (f) indicate that removing spatio‑temporal semantic feedback modules in certain layers degrades target feature extraction, further underscoring the necessity of the complete feedback framework.

\subsubsection{Sparse Grouping Method}
To evaluate the effectiveness of our sparse grouping strategy within the SSM, we redesign several alternative methods for construction of sparse frames, which are summarized as follows. Among them, the last one is the method we ultimately adopt for grouping frames.
\begin{itemize}
	\item{\textbf{Group-Rand-Seq}: We randomly segment the input continuous frames into contiguous segments and perform alignment only between consecutive frames within each segment.}
	\item{\textbf{Group-Rand-Step-Fixed}: We randomly determine a global sampling interval and uniformly adopt this interval across the network to achieve skip-frame alignment.}
	\item{\textbf{Group-Rand-Step-Rand}: We assign a random sampling interval to each layer, enabling skip-frame alignment with layer-wise randomness.}
	\item{\textbf{Group-Fixed-Step-T2}: We manually preset the sampling interval to $T=2$ as our final configuration to achieve skip-frame alignment.}
\end{itemize}

As shown in Table~\ref{tab:group_type_table}, among all variants, Group-Fixed-Step-T2 achieves the best overall performance, attaining the highest $mIoU$ (48.97), $F_1$ (65.74), $P_d$ (96.34), and $AUC$ (98.09), while maintaining a competitive false alarm rate (1.19). Group-Rand-Step-Fixed delivers strong second-best results in $P_d$ (92.35), $F_a$ (1.18), and $AUC$ (96.10), demonstrating the effectiveness of a globally consistent sampling interval. Notably, Group-Rand-Step-Rand achieves slightly higher $mIoU$ and $F_1$ than Group-Rand-Step-Fixed, but lags in $P_d$ and $AUC$, indicating that layer-wise random sampling may disrupt stable temporal propagation. In contrast, Group-Rand-Seq shows the weakest performance across all metrics, suggesting that purely contiguous segmentation without skip-frame propagation limits the model’s ability to capture long-range dependencies. These results validate our final design choice of fixed-step sampling as the optimal configuration for balancing temporal coverage and propagation effectiveness.
\begin{table}[htbp]
	\centering
	\renewcommand{\arraystretch}{1.3}
	\setlength{\tabcolsep}{4.3pt} 
	\caption{Ablation results of $mIoU\left(\times10^{-2}\right)$, $F_1\left(\times10^{-2}\right)$, $P_d\left(\times10^{-2}\right)$, $F_a\left(\times10^{-6}\right)$ and $AUC\left(\times10^{-2}\right)$ are achieved by different frame alignment grouping strategies on the dataset IRSatvideo-LEO\cite{ying2025infrared}. The rows containing our final proposed framework are distinguished with gray background highlighting. The best-performing results are bolded, while the second-best results are underlined.}
	\label{tab:group_type_table}
	\begin{tabular}{l c c c c c}
		\toprule
		\textbf{Framework} & $\bm{mIoU}\uparrow$ & $\bm{F_1}\uparrow$ & $\bm{P_d}\uparrow$ & $\bm{F_a}\downarrow$ & $\bm{AUC}\uparrow$ \\
		\midrule
		Group-Rand-Seq & 32.19 & 48.70 & 88.16 & 5.59 & 94.00 \\
		Group-Rand-Step-Fixed & 43.49 & 60.62& \underline{92.35} & \textbf{1.18} & \underline{96.10} \\
		Group-Rand-Step-Rand & \underline{43.78} & \underline{60.90} & 88.63 & 1.23 & 94.26 \\
		\rowcolor{gray!20}
		\rowcolor{gray!20}
		Group-Fixed-Step-T2 & \textbf{48.97} & \textbf{65.74} & \textbf{96.34} & \underline{1.19} & \textbf{98.09} \\
		\bottomrule
	\end{tabular}
\end{table}

\subsubsection{Sampling Interval in SSM}
In this experiment, we compare the impact of different temporal sampling intervals of the SSM module on performance. For clarity, we use T1 to denote the baseline without temporal sampling, and T2, T3, T4 as shorthands for FeedbackSTS-Det-T2, -T3, and -T4, with sparse frame sampling intervals of 2, 3, and 4, respectively. The computational efficiency of the model under different settings is summarized in Table \ref{tab:FPS_compare}, which reports FLOPs and FPS for various sequence lengths  $L$ and temporal sampling intervals $T$ based on an input image scale of $256\times256$. The results demonstrate that for a fixed $T$, longer sequences $L$ lead to higher FLOPs and generally lower FPS.  Conversely, increasing the temporal sampling interval $T$ consistently reduces computational cost and improves inference speed across all sequence lengths. For instance, when the sequence length is set to $L=9$, the computational cost measured in FLOPs decreases from 68.42G for $T=1$ to 52.08G for $T=4$, while the inference speed (FPS) increases from 7.5 to 13.7, representing a substantial improvement of 82.6\% in processing efficiency.This analysis confirms that larger temporal sampling intervals enhance computational efficiency, especially for long sequences.

\begin{table}[!t]
	\caption{Ablation study of computational efficiency (FLOPs and FPS) for varying sequence lengths $L$ and sparse frame sampling intervals $T$. Both of FLOPs and FPS are calculated based on an input image scale of $256\times256$.}
	\label{tab:FPS_compare}
	\centering
	\renewcommand{\arraystretch}{1.5}
	\setlength{\tabcolsep}{2.3pt} 
	\scriptsize
	\begin{tabular}{c|cc|cc|cc|cc}
		\hline
		\multirow{2}{*}{\centering $\bm{L}$} & \multicolumn{2}{c|}{$\bm{T=1}$} & \multicolumn{2}{c|}{$\bm{T=2}$} & \multicolumn{2}{c|}{$\bm{T=3}$} & \multicolumn{2}{c}{$\bm{T=4}$} \\ \cline{2-9}
		~ & \textbf{FLOPs}$\downarrow$ & \textbf{FPS}$\uparrow$ & \textbf{FLOPs}$\downarrow$ & \textbf{FPS}$\uparrow$ & \textbf{FLOPs}$\downarrow$ & \textbf{FPS}$\uparrow$ & \textbf{FLOPs}$\downarrow$ & \textbf{FPS}$\uparrow$\\ \hline
		5 & 35.59G & 8.4 & 30.14G & 10.7 & 24.70G & 14.2 & 19.25G & 15.1 \\ 
		7 & 52.00G & 8.0 & 46.56G & 9.8 & 41.11G & 11.2 & 35.67G & 15.2 \\
		9 & 68.42G & 7.5 & 62.97G & 8.9 & 57.53G & 10.5 & 52.08G & 13.7 \\
		11 & 84.83G & 7.6 & 79.39G & 8.3 & 73.94G & 8.9 & 68.50G & 9.7 \\
		13 & 101.245G & 7.4 & 95.80G & 7.6 & 90.36G & 9.1 & 84.91G & 10.7 \\
		15 & 117.66G & 7.7 & 112.214G & 7.8 & 106.77G & 8.9 & 101.32G & 9.8 \\
		\hline
	\end{tabular}
\end{table}

\begin{figure}[!t]
	\centering
	\subfloat[]{
		\includegraphics[width=3.5in]{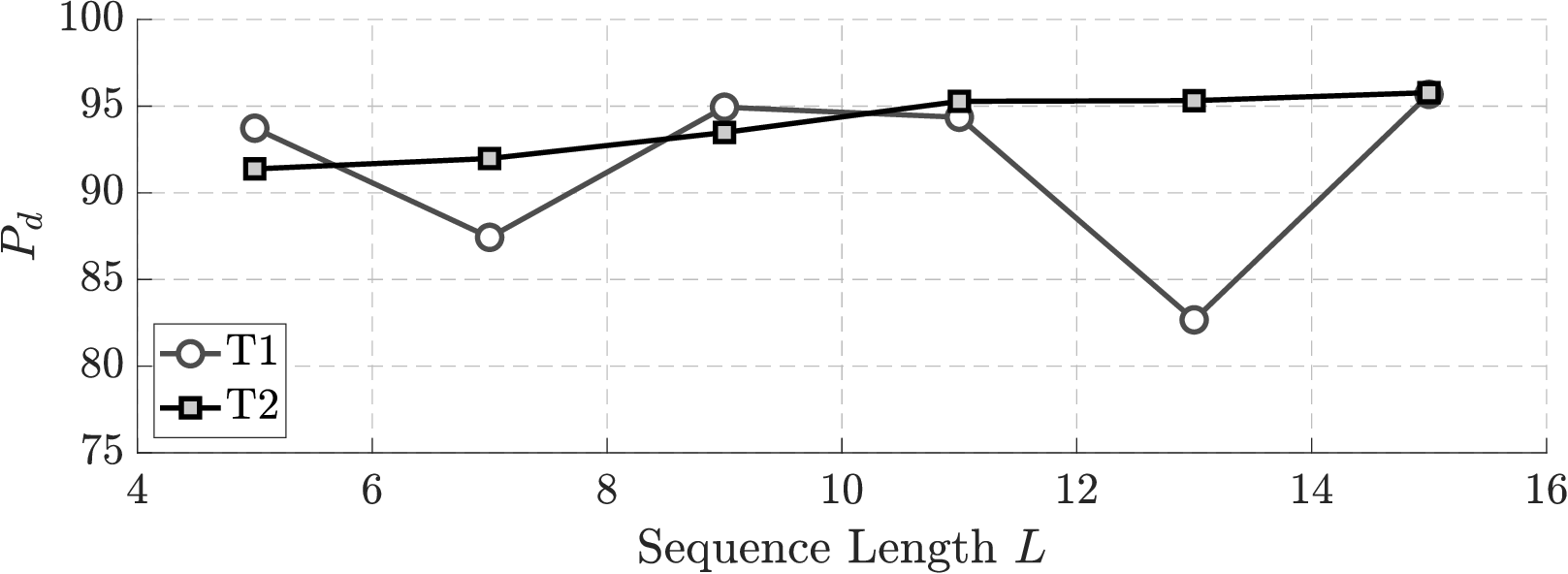}
		\label{fig:stable_pd}
	}
	\vfil
	\subfloat[]{
		\includegraphics[width=3.5in]{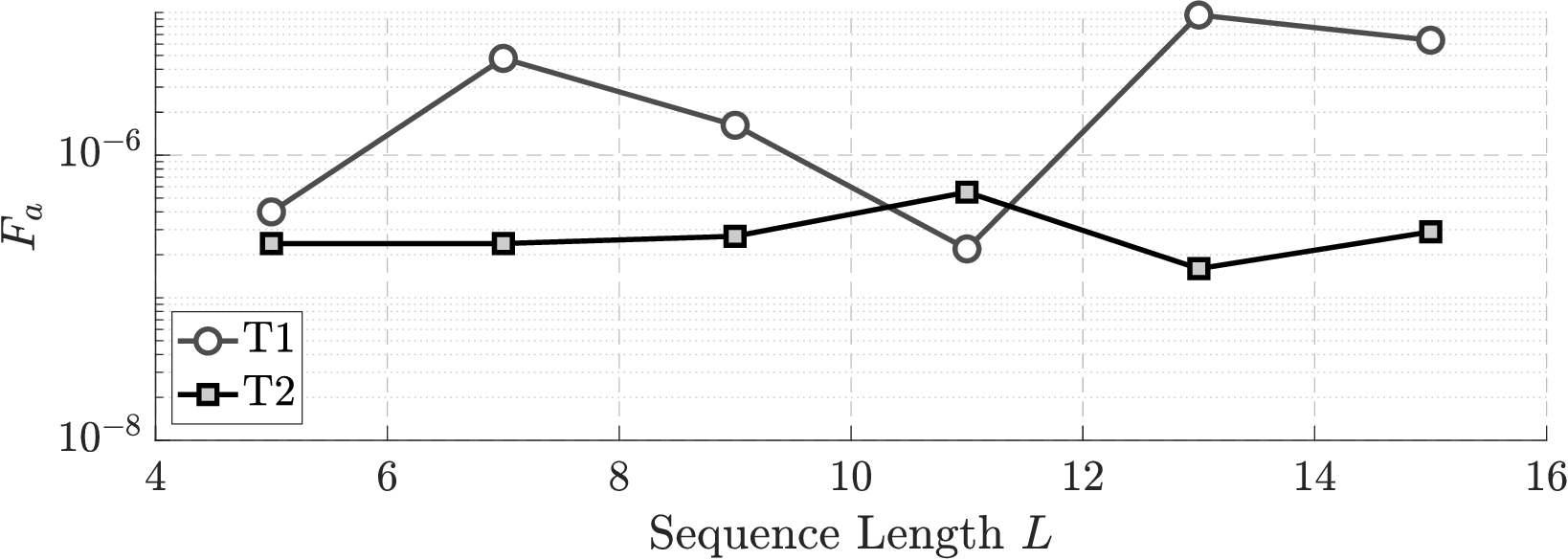}
		\label{fig:stable_fa}
	}
	\caption{Ablation study for the IRSatVideo-LEO \cite{ying2025infrared} under no temporal sampling and T2 conditions. (a) $P_d$ vs. $L$. (b) $F_a$ vs. $L$.}
	\label{fig:stable}
\end{figure}

To evaluate the efficacy of the temporal sampling mechanism, we conduct a comparative analysis between the T2 condition (with a temporal sampling interval of length 2) and the T1 condition (without temporal sampling) on the IRSatVideo-LEO \cite{ying2025infrared}, using $P_d$ and $F_a$ as key metrics. As shown in Fig.~\ref{fig:stable} (a), the $P_d$ for T2 increases monotonically with the sequence length $L$, whereas the $P_d$ for T1 exhibits significant and unstable fluctuations. From Fig.~\ref{fig:stable} (b), it is observed that the $F_a$ for T2 remains consistently stable within a narrow range as $L$ increases. In contrast, the $F_a$ for T1 varies drastically and is markedly higher than that of T2. In summary, the sparse frame sampling module significantly enhances target detection capability while effectively suppressing false alarms over long sequences, demonstrating its robust ability to propagate inter-frame features compared to the baseline T1 condition.

\begin{figure}
	\centering
	\includegraphics[width=3.5in]{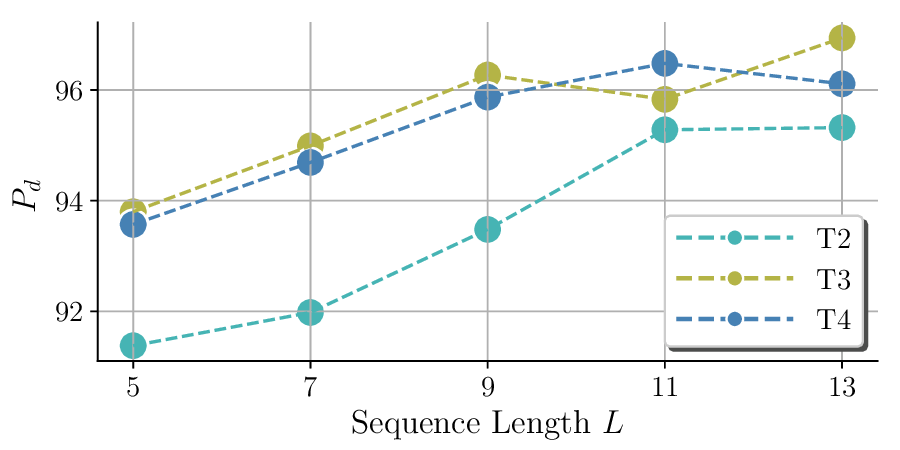}
	\caption{Ablation results of Sequence Length Effect on $P_d$ under different sampling intervals for the IRSatVideo-LEO \cite{ying2025infrared}.}
	\label{fig:pd_sequence}
\end{figure}

To investigate the capability of different sparse frame sampling intervals in extracting targets under varying input window sizes, we evaluate the effect of the input sequence length $L$ on the detection probability $P_d$ cross different sampling intervals using the IRSatVideo-LEO \cite{ying2025infrared}, as shown in Fig.~\ref{fig:pd_sequence}. It could be observed that $P_d$ increases with $L$ for all sampling intervals, validating that the sparse frame sampling module successfully enhances inter-frame dependencies in long sequences. Moreover, a longer sampling interval is generally associated with improved performance at an equivalent sequence length. An exception is noted at $L=11$, where T4 underperforms T3; however, T4 demonstrates superior performance at $L=13$.

In summary, the experimental results suggest that increasing the sampling interval moderately reduces the network’s computational cost while improving its temporal propagation capability across long frame sequences. However, an overly large sampling interval $T$ can compromise detection accuracy. Our experimental results confirm that choosing $T=2$, $3$ or $4$ reduces computational cost to a certain extent while preserving or even improving the detection performance of the sequence.

\FloatBarrier
\section{Conclusion} \label{sec:conclusion}
This paper presents a novel sparse frames-based spatio-temporal feedback network, i.e., FeedbackSTS Detector, for moving infrared small target detection. Its core is a closed-loop semantic feedback strategy with forward and backward refinement modules that work jointly, enabling information exchange between consecutive frames, thereby improving detection accuracy and reducing false alarms. More importantly, an embedded sparse semantic module (SSM) performs structured sparse temporal modeling to capture long-range dependencies with low computational cost. Extensive experiments on many widely adopted multi-frame infrared small target datasets demonstrate the generalization ability and scene adaptability of our proposed FeedbackSTS-Det. The limitation of our work is that experiments are conducted on only a limited number of scenarios, which may not fully represent the diversity of real-world infrared environments. Therefore, the model's robustness may be insufficient when dealing with extremely dim targets, strong clutter, or fast-moving objects, which could affect the generalizability of the results. In future work, we will optimize the network architecture and evaluate it on a more diverse range of infrared data scenarios, aiming to progressively improve the model's generalization capability.

\bibliographystyle{IEEEtran}
\bibliography{reference}

\section{Biography Section}
\vspace{1pt}
\begin{IEEEbiography}[{\includegraphics[width=1in,height=1.25in,clip,keepaspectratio]{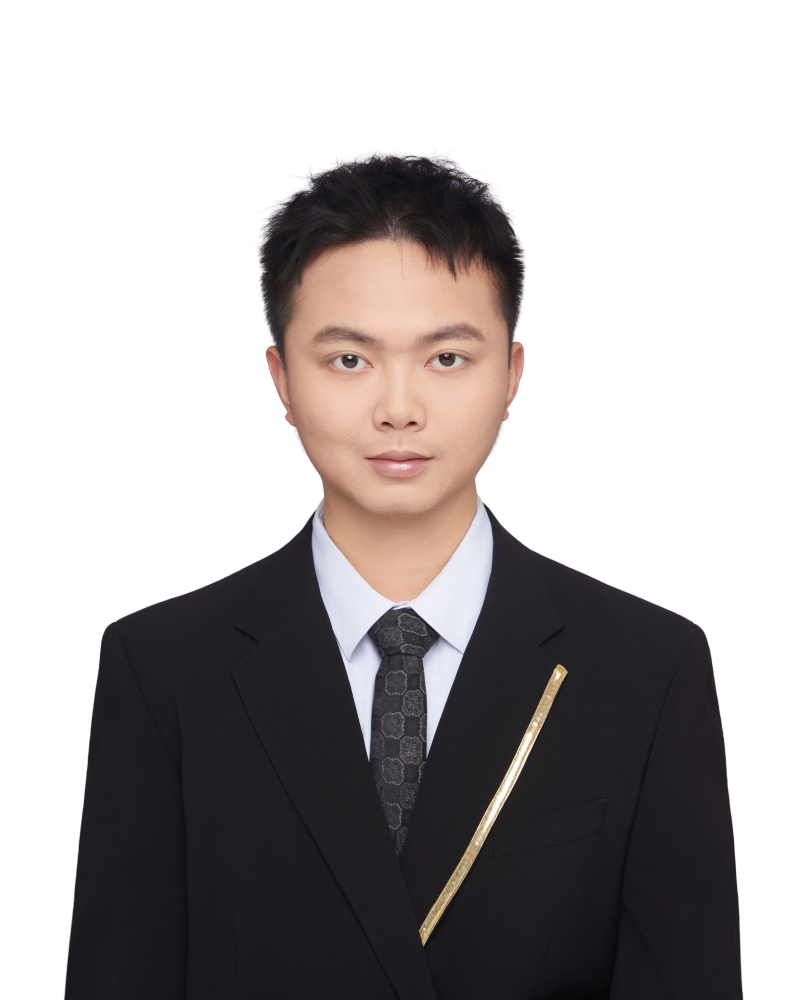}}]{Yian Huang}
received the B.S. degree in communication engineering from Sun Yat-sen University (SYSU), Guangzhou, China, in 2022. He is currently pursuing the Ph.D. degree with the School of Information and Communication Engineering from University of Electronic Science and Technology of China (UESTC), Chengdu, China. His research interests include image processing, computer vision and infrared target recognition.
\end{IEEEbiography}
\vspace{1pt}
\begin{IEEEbiography}[{\includegraphics[width=1in,height=1.25in,clip,keepaspectratio]{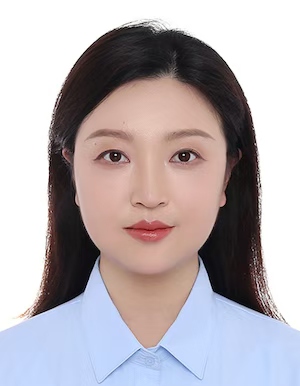}}]{Qing Qin} received her Ph.D. degree in arts from the Daejin University, Gyeonggi-do, Korea, in 2024. She is currently a lecturer with the Communication University of Zhejiang, Hangzhou. Her research interests include intelligent image production , digital video processing, and future advanced imaging.
\end{IEEEbiography}
\vspace{1pt}
\begin{IEEEbiography}[{\includegraphics[width=1in,height=1.25in,clip,keepaspectratio]{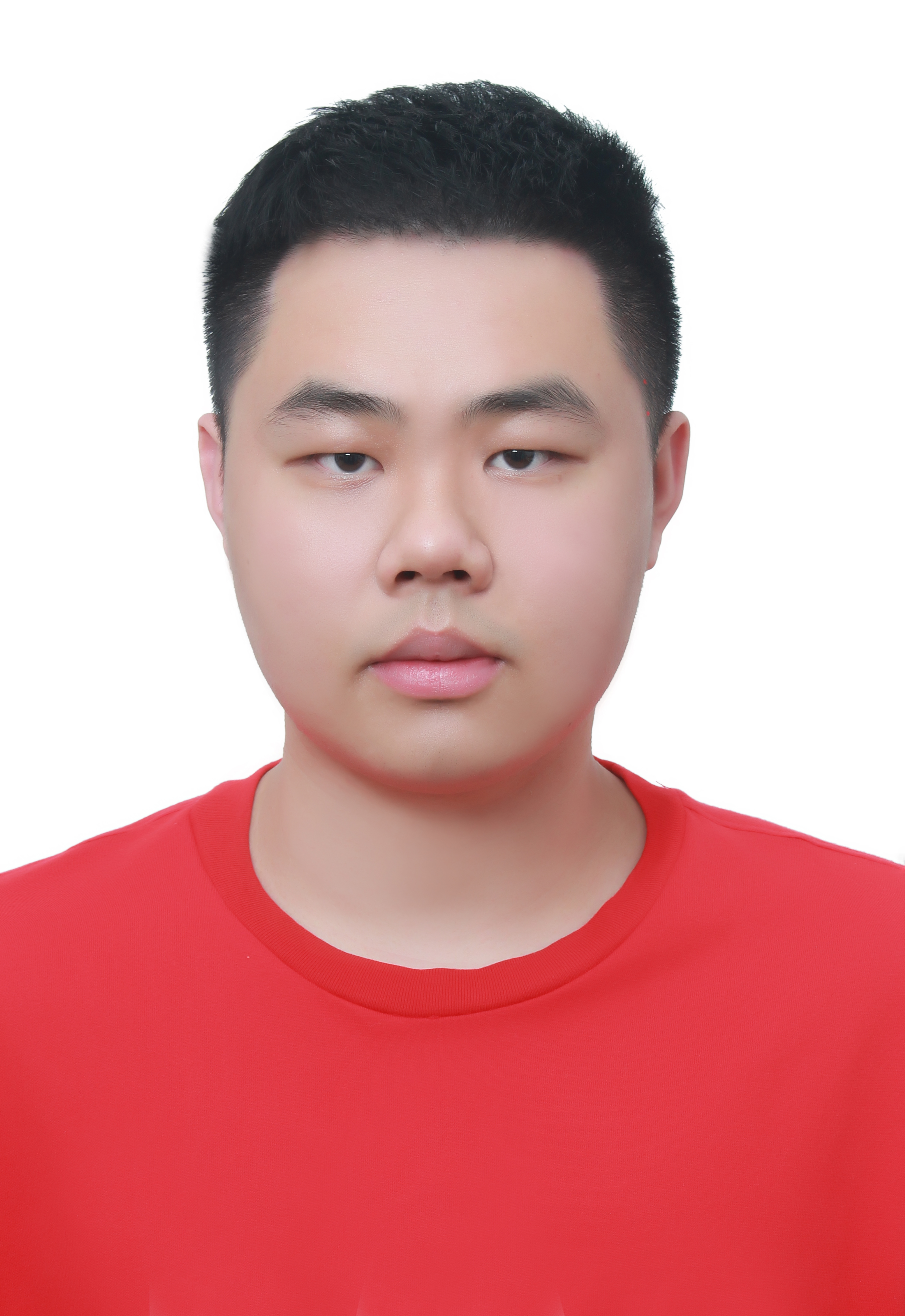}}]{Aji Mao}
received the B.S. degree in electronic information engineering from China University of Mining and Technology (CUMT), Xuzhou, China, in 2021. He is currently pursuing the M.S. degree with the School of Information and Communication Engineering from University of Electronic Science and Technology of China (UESTC), Chengdu, China. His research interests include computer vision, large language model and infrared target recognition.
\end{IEEEbiography}
\vspace{1pt}
\begin{IEEEbiography}[{\includegraphics[width=1in,height=1.25in,clip,keepaspectratio]{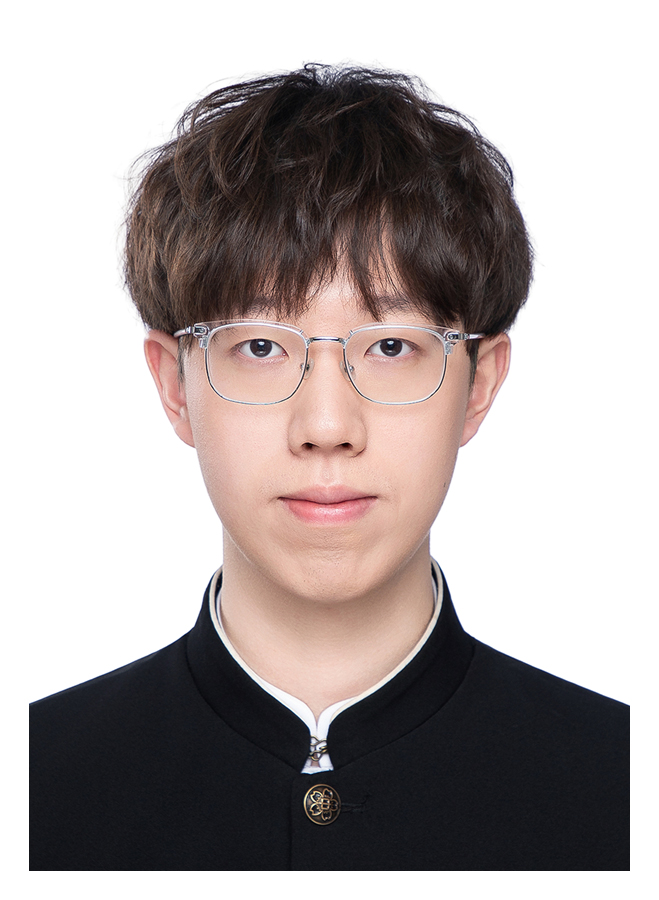}}]{Xiangyu Qiu} received his B.E. degree from the school of Information and Communication Engineering, University of Electronic Science and Technology of China, in 2024. He is pursuing an M.E. degree in School of Information and Communication Engineering, University of Electronic Science and Technology of China. His re-search interests include image processing, computer vision, and infrared small target detection.
\end{IEEEbiography}
\vspace{1pt}
\begin{IEEEbiography}[{\includegraphics[width=1in,height=1.25in,clip,keepaspectratio]{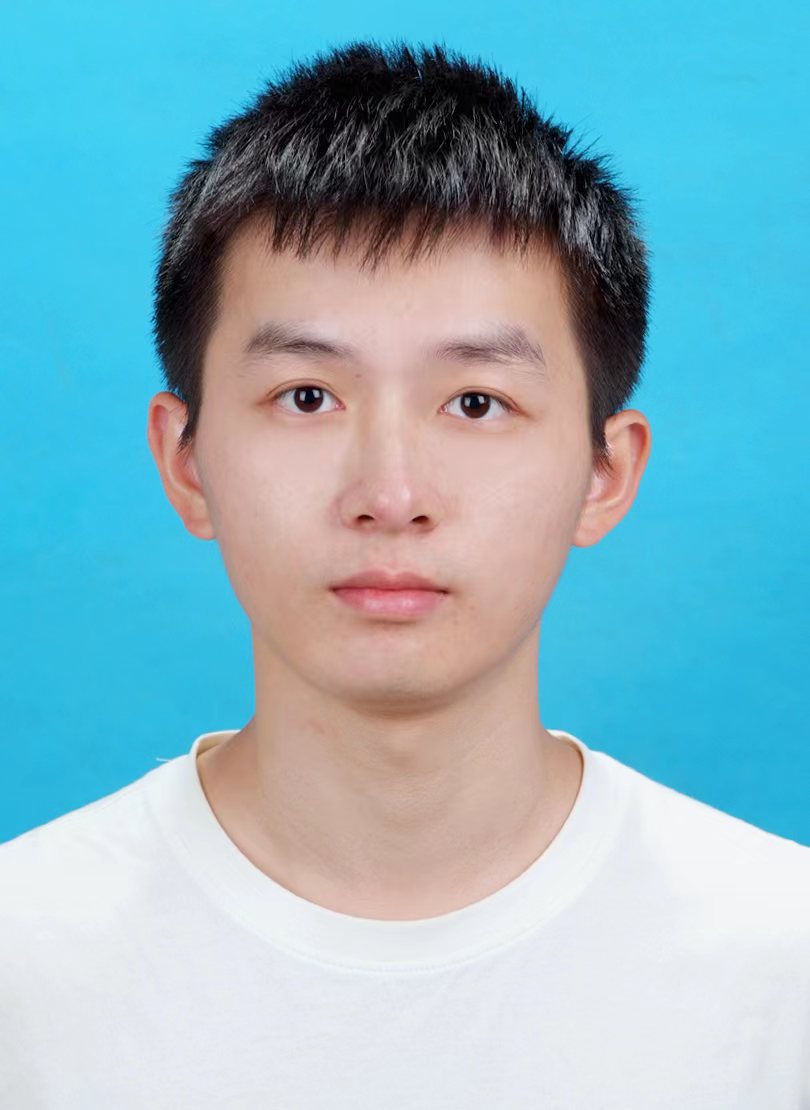}}]{Liang Xu} received the B.E. degree in the School of Information and Communication Engineering from University of Electronic Science and Technology of China (UESTC), Chengdu, China, in 2024. He is currently pursuing the M.S. degree with the School of Information and Communication Engineering from UESTC, Chengdu, China. His research interests include image processing, computer vision and infrared target detection.
\end{IEEEbiography}
\vspace{1pt}
\begin{IEEEbiography}[{\includegraphics[width=1in,height=1.25in,clip,keepaspectratio]{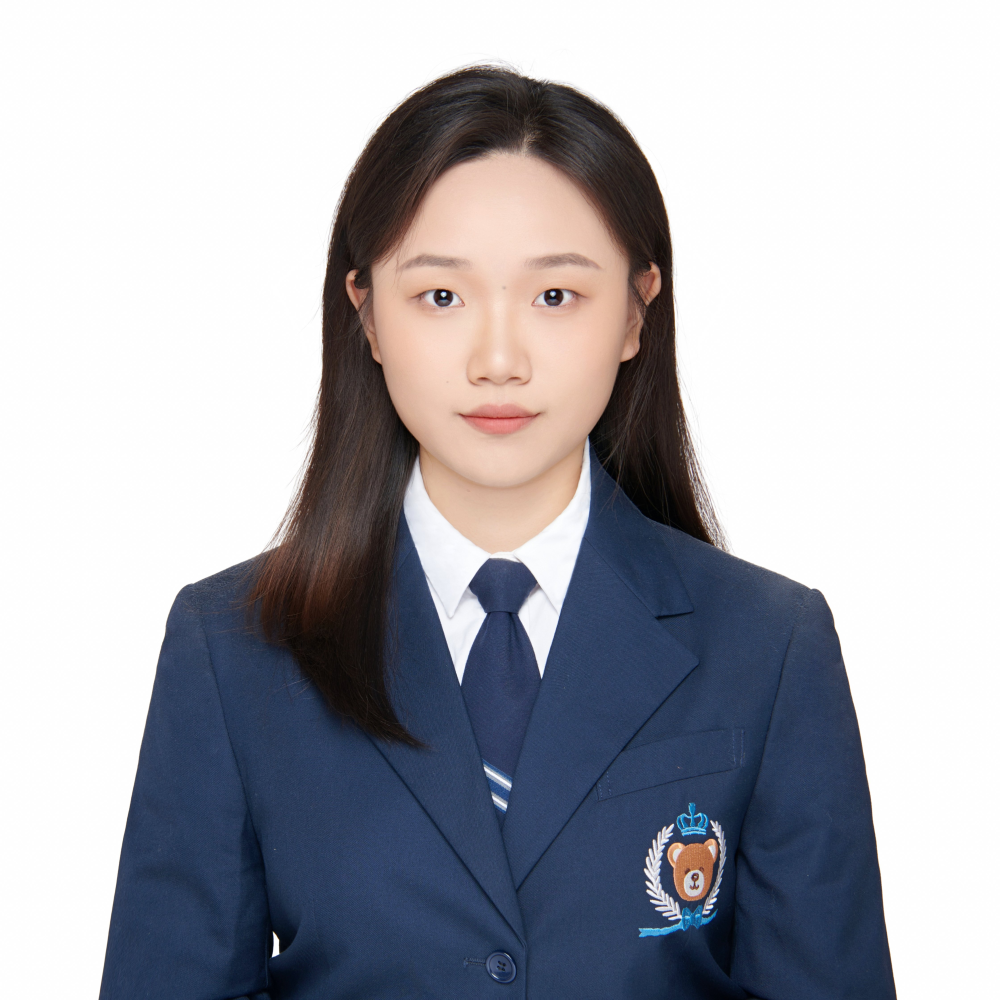}}]{Xian Zhang} received the B.S. degree in Information and Communication Engineering from University of Electronic Science and Technology of China (UESTC), Chengdu, China, in 2024. She is currently pursuing the M.S. degree with the School of Information and Communication Engineering, University of Electronic Science and Technology of China (UESTC), Chengdu, China. Her research interests include image processing and machine learning.
	
\end{IEEEbiography}
\vspace{1pt}
\begin{IEEEbiography}[{\includegraphics[width=1in,height=1.25in,clip,keepaspectratio]{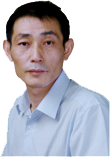}}]{Zhenming Peng} (Senior Member, IEEE) received his Ph.D. degree in geodetection and information technology from the Chengdu University of Technology, Chengdu, China, in 2001. From 2001 to 2003, he was a Post-Doctoral Researcher with the Institute of Optics and Electronics, Chinese Academy of Sciences, Chengdu, China. He is currently a Professor with the University of Electronic Science and Technology of China, Chengdu. His research interests include image processing, machine learning, objects detection and remote sensing applications.
\end{IEEEbiography}

\vfill

\end{document}